\pdfoutput=1

\documentclass[11pt]{article}

\usepackage[]{acl}

\usepackage{times}
\usepackage{latexsym}

\usepackage[T1]{fontenc}

\usepackage[utf8]{inputenc}

\usepackage{microtype}

\usepackage{inconsolata}
\usepackage{graphicx}
\usepackage{hyperref}
\usepackage{pgffor} 
\usepackage{multirow}
\usepackage{tikz}
\usepackage{pgfplots}
\usetikzlibrary{patterns}
\usepackage{booktabs}
\usepackage{makecell}
\usepackage{xspace}
\usepackage{balance}
\usepackage{amsmath}
\usepackage{verbatim}

\usepackage[textsize=scriptsize]{todonotes}

\newcommand{\eat}[1]{}

\newcommand{\task}[0]{EEL\xspace}
\newcommand{\method}[0]{CGEL\xspace}

\title{Collaborative Causal Graph Generation for Event Reasoning}
\title{Causal Graph Generation for Event Reasoning using Collaborative Agents}
\title{Generating Causal Event Graphs using Collaborative Agents}
\title{Generating Causal Event Graphs using Semantic Relations Experts}
\title{Event Reasoning with Causal Graphs from Semantic Relations Experts}
\title{Semantic Relation Experts for Causal Graph Generation in Event Reasoning}
\title{Semantic Relation Experts for Causal Graph based Event Reasoning}
\title{Causal Graph based Event Reasoning using Semantic Relation Experts}

\author{Mahnaz Koupaee$^1$, Xueying Bai$^1$, 
Mudan Chen$^2$, \\ \textbf{Greg Durrett$^3$, Nathanael Chambers$^4$, Niranjan Balasubramanian$^1$ }\\
\\
  $^1$Stony Brook University,
  $^2$University of Pennsylvania,
  \\
  $^3$The University of Texas at Austin,
  $^4$United States Naval Academy\\
  $^1$\texttt{\{mkoupaee,xubai,niranjan\}@cs.stonybrook.edu}\\
  $^2$\texttt{mdchen@seas.upenn.edu},
  $^3$\texttt{gdurrett@cs.utexas.edu},
  $^4$\texttt{nchamber@usna.edu}\\}

\begin{document}
\maketitle
\begin{abstract}
\color{black}{
Understanding how events in a scenario \emph{causally} connect with each other is important for effectively modeling and reasoning about events.
But event reasoning remains a difficult challenge, and despite recent advances, Large Language Models (LLMs) still struggle to accurately identify causal connections between events.
This struggle leads to poor performance on deeper reasoning tasks like event forecasting and timeline understanding.
To address this challenge, we investigate the generation of causal event graphs (e.g., A enables B) as a parallel mechanism to help LLMs explicitly represent causality during inference.
This paper evaluates both how to generate correct graphs as well as how graphs can assist reasoning.
We propose a collaborative approach to causal graph generation where we use LLMs to simulate experts that focus on specific semantic relations.
The experts engage in multiple rounds of discussions which are then consolidated by a final expert.
Then, to demonstrate the utility of causal graphs, we use them on multiple downstream applications, and also introduce a new explainable event prediction task that requires a causal chain of events in the explanation.
These explanations are more informative and coherent than baseline generations.
Finally, our overall approach not finetuned on any downstream task, achieves competitive results with state-of-the-art models on both forecasting and next event prediction tasks.}

\end{abstract}

\section{Introduction}


Event understanding and reasoning has a rich and long history ranging from early works that modeled events through manually specified schemas \cite{schank-abelson-75, mooney1985learning}, statistical event co-occurrence models~\cite{manshadi2008learning, chambers2013event, balasubramanian2013generating}, event language models ~\cite{modi2016event, pichotta2016learning,weber2018event,weber2018hierarchical,koupaee2021don}, and controllable generation with larger language models~\cite{gunjal2023drafting,tang2023harvesting,dror2023zero}. While recent work has advanced event reasoning capabilities overall, particularly in event prediction, most work has relied primarily on distributional relations between events captured by co-occurrences. Some notable exceptions have sought to model deeper connections targeting specific semantic relations between events~\cite{rezaee2021event,rezaee-ferraro-2023-revup,koupaee-etal-2023-modeling,li-etal-2021-future}. A comprehensive understanding of events, however, requires understanding the rich set of logical connections that explain how the set of events unfold in a scenario \cite{wanzare2016crowdsourced,li2020connecting,li2021future,wang2022schema}. 
\begin{figure}
    \centering
    \includegraphics[width=0.95\linewidth]{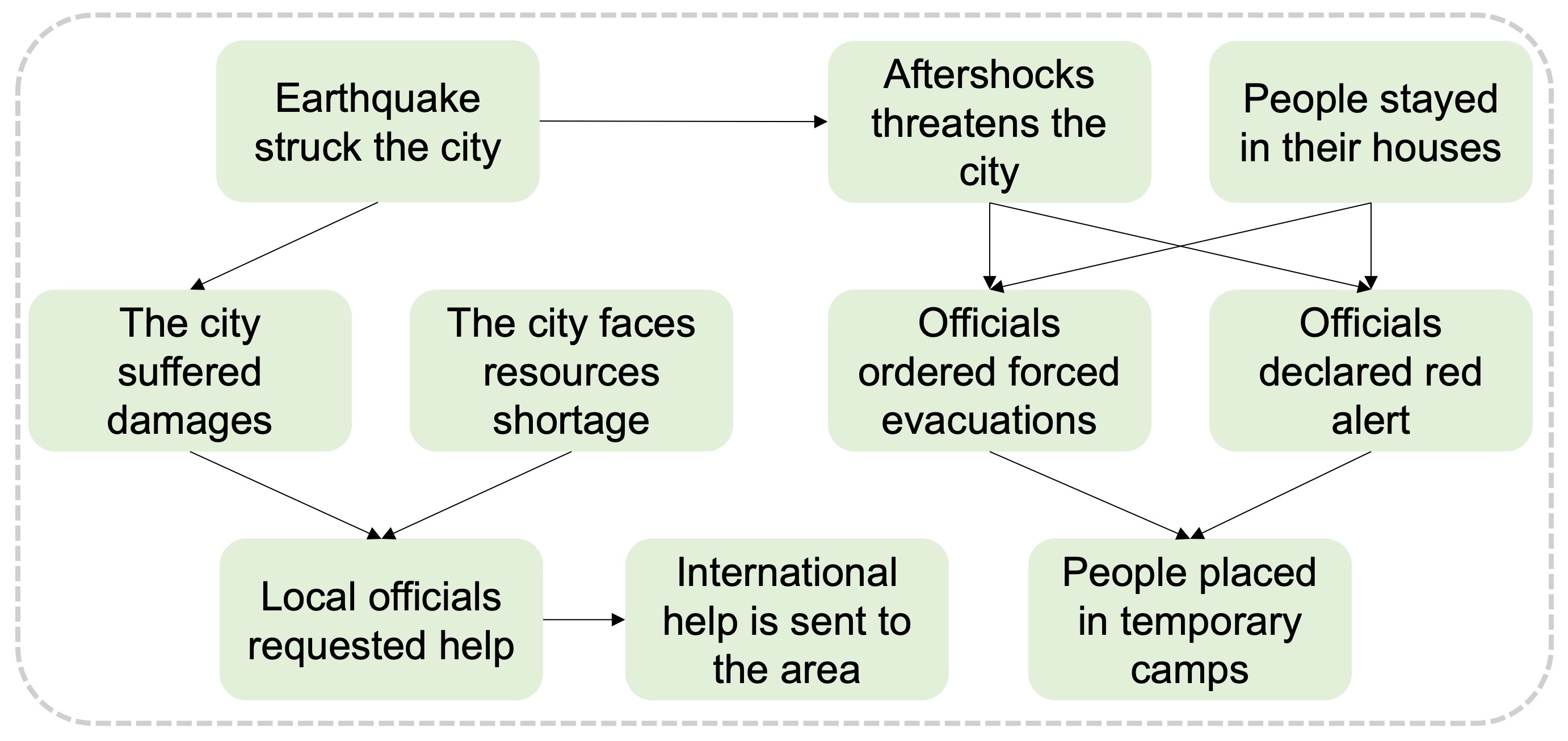}
    \caption{\color{black}{A causal graph consisting of events as nodes and causal relations as directed edges.}}
    \label{fig:eelp-task-overview}
\end{figure}
In this work, we target global causal connections between events as a means to understanding and reasoning about events, targeting applications such as event likelihood prediction \cite{modi2016event, pichotta2016learning,weber2018event} and event forecasting \cite{jin2021forecastqa}. However, identifying causal relations between events themselves remains a challenging task. Indeed, previous results highlight the difficulties in identifying causal relations in pairwise settings \cite{romanou2023crab,sun2023event}, and as we show later even state-of-the-art LLMs struggle to get high accuracies for this task under standard in-context learning settings.

{\color{black}Identifying causal relations is challenging because it requires considering how the events in question are embedded in the larger global context of the other events in the scenario. 
For instance, using Figure 1 as an example, earthquakes are fairly common and not always catastrophic, so local officials don't always request help. However, if a city "faced resources shortage" \emph{and} the "city suffers damage", these two events combine to enable "local officials requested help". Without such logical connections explicitly spelled out, an LLM might miss this subtle reasoning.}

{\color{black}Having established the need for causal graphs, can LLMs generate them?
To answer this question, we introduce a novel causal graph construction method that produces global causal graphs using a set of collaborative agents, each specializing over a specific aspect of causality. 
Causal relations can be supported or refuted when considered through different aspects, especially when evidence from each of the aspects themselves are inferred by a model. For example, an incorrectly inferred temporal relation might in turn erroneously suggest a causal direction, whereas considering a common-sense aspect might provide the needed correction for both the temporal and causal relation. These suggest a collaborative agentic approach with rounds of communication to arrive at a consensus around the global graph structure.
To achieve this, we introduce four relation experts: \textit{temporal}, \textit{discourse}, \textit{pre-condition} and \textit{commonsense} who are responsible for identifying causal relations while focusing on their own expertise. 
These experts engage in \emph{conversations} to learn the views of the other experts and \textit{collaborate} to arrive at a consensus on the global causal structure.}
Our intrinsic evaluation shows the effectiveness of both \textit{relation experts} and \textit{collaboration} in generating more accurate causal graphs compared to other LLM-based variants. 
For our extrinsic evaluation, we first pose a new explainable event likelihood prediction task (\task) that requires models to provide a causal chain of events to support event likelihood predictions. 
Then we show that our approach of reasoning over causal graphs yields significantly better predictions and explanations compared to few-shot GPT-4 baselines and with Reflexion \cite{shinn2023reflexion}. We further demonstrate the downstream utility of these causal models for a forecasting task (ForecastQA) and a variant of a narrative cloze task. 

{\color{black}
In summary, this paper makes the following main contributions: (1) introduces a collaborative approach with the help of relation experts to generate causal graph of events that can be used to drive predictions and produce explanations, (2) introduces a new explainable event likelihood prediction task (\task) and (3) provides an empirical evaluation demonstrating the utility of our collaborative approach to generate causal graphs and effectiveness of such graphs on a diverse set of applications\footnote{Code available at \href{https://github.com/StonyBrookNLP/causal-graphs}{https://github.com/StonyBrookNLP/causal-graphs}}.}

\section{Collaborative Causal Graph Generation}
\label{sec:celp}
\begin{figure*}
    \centering
    \includegraphics[width=1\linewidth]{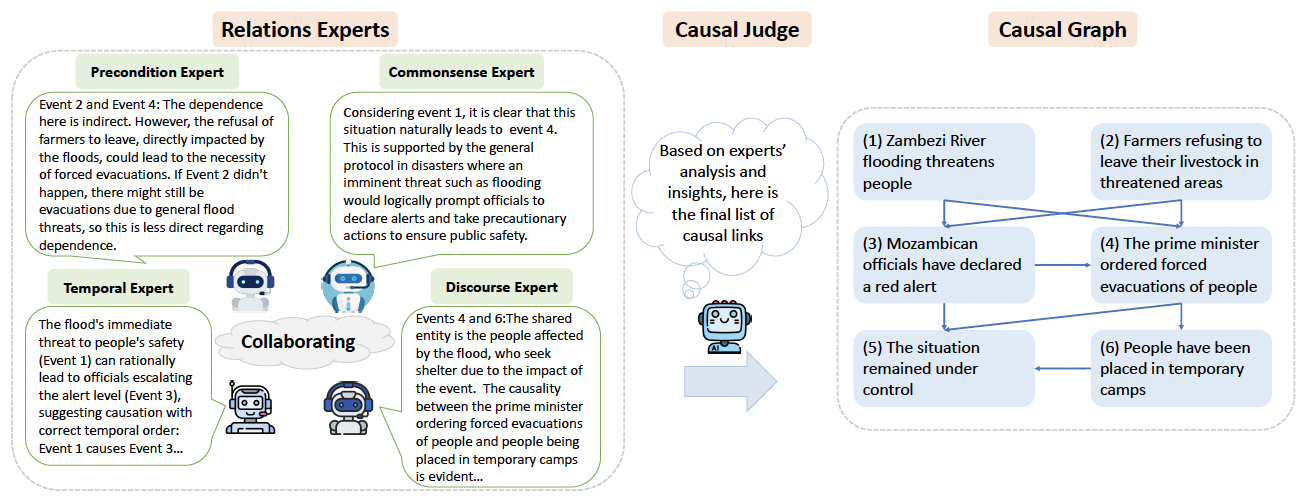}
    \caption{Collaborative causal graph generation with relation experts. Initially, the experts start communicating with each other, presenting their findings on causality based on their expertise. After a certain number of communications, a judge would compile the finalized list of causal link of a given scenario.}
    \label{fig:overview_causal_graph_gen}
\end{figure*}
Identifying causal relations has proven to be a difficult task specially for news domain. Even SOTA LLMs with their strong reasoning capabilities are still far from perfect on identifying causal relations \cite{romanou2023crab}.
Many event reasoning tasks require producing a global causal structure (a causal backbone) that ties together the set of events\footnote{{\color{black}We rely on the same notion of event used in prior work and the specific datasets we use in our experiments. Specifically, an \textit{event} is a representation of an action that would result in a status change of the world and it consists of predicates (verbs) and participants (arguments) such as subjects and objects.}}.
As we argued earlier, different semantic relations can contribute to causality detection.
For instance, a temporal relation between an event pair is a necessary (but not a sufficient) condition for causality. Using this type of explicit information about other semantic relations can  help reduce errors in identifying causality.

 A direct approach to achieve these goals is to directly prompt an LLM to consider all of these aspects when making its decisions about causality. However, reasoning about each aspect is in itself difficult and can have errors, which are also best resolved by inspecting the arguments and reassessing inferences. It is challenging even for SOTA LLMs to assess all of these complex threads of information in \emph{one go}. Thus, we need an approach which fosters a ``separation of concerns'' by simulating different experts who argue, reflect, and revise their reasoning using information from other experts. 

To this end, we propose a collaborative approach shown in Figure \ref{fig:overview_causal_graph_gen} in which relation experts (agents), each with their unique perspective, work collaboratively to produce a causal graph that connects the set of events in a discourse. In this section, we first describe the relation experts we use and the collaboration scheme we developed in detail. \autoref{sec:intrinsic-eval} demonstrates the effectiveness of our approach for causal graph generation, and \autoref{sec:extrinsic-eval} highlights the utility of the causal graphs thus generated in multiple event reasoning tasks.


\subsection{Relation Experts}\label{sec:relations-epxerts}

Our main idea is to have multiple experts each tasked with identifying causal relations but are instructed to focus on a specific semantic relation when making their causal decision. We use LLMs to mimic four such experts: temporal, discourse, dependence and commonsense relations. 
Each expert is asked to identify the causal relations of the given context but they are assigned specific dimensions to focus on.
We describe what each of these experts is supposed to do next.

\noindent\textbf{Temporal Expert:}
If $e_i$ precedes $e_j$ , then $(e_i, e_j)$ has a temporal relation. A temporal relation is a necessary condition for causality as a cause always precedes its effect.
In a causal relation, the cause always precedes the effect and therefore a temporal link between the events can be seen as a prerequisite for causality. 
The temporal expert is instructed to identify all event pairs that share a temporal relation, filter out the ones with no clear temporal relation, looks for causal links among the ones with temporal relations and also helps other experts by making the search space smaller.

\noindent\textbf{Discourse Expert:}
If there exists an entity which participates in both $e_i$  and $e_j$ , then $(e_i, e_j)$ are said to have a discourse relation. 
Events sharing entities are potential candidates to have a causal relation. The rationale here is that the action on an entity in one event can lead to subsequent events with the same entity. Though not always the case, analyzing events from this dimension can also help scope the causality detection problem to a manageable set. The discourse expert thus tries to identify causal links from the perspective of discourse relations.

\noindent\textbf{Conditional Expert:}
Given a pair of events $(e_i, e_j)$ we assert $e_i$ is a precondition event for $e_j$ if $e_j$ likely would not have occurred without $e_i$ \cite{kwon2020modeling,han2021ester}.
The conditional expert is responsible for considering alternative scenarios and see whether removing an event can have an impact on other events and thus identify potential candidates that can be causally connected with respect to this aspect. 

\noindent\textbf{Commonsense Expert:} 
In some cases, the causal relation between $e_i$ and $e_j$ is mediated by relations to some implicit (i.e. unstated) information. 
Identifying such missing pieces of information that are not explicitly mentioned, can help determine whether there is a direct or indirect causal relation between events. Therefore, the commonsense expert is tasked with focusing on the implicit knowledge that connects the explicitly mentioned events in the context to determine causality.

\subsection{Collaboration Scheme}
Experts are initialized with specific roles and tasks as described above. They generate their first set of responses independently and then conduct multiple rounds of discussions with each other. At each round, each expert gets the responses from all other experts, analyzes all responses and generates a revised list of potential causal links along with their reasoning.
The discussion is aimed at providing reasoning as to why each expert thinks an event pair is causal (or not) and they try to collaborate with each other towards finalizing the list of all possible causal links.
They continue the discussion for a specified number of rounds or until all agree on the causal links (Prompts in Appendix \ref{causal-graph-prompts}). 

\subsection{Causality Judge}
Once the discussion is concluded, regardless of how long it went on, the experts may have reached consensus or might still have some disagreements left. To resolve any remaining disagreements and to compile a final list of responses, we ask an LLM to act as a final arbiter, a causality judge, who would summarize the other experts contributions to the discussion and finalize the list of causal links. 
\paragraph{On importance of context}
Transitivity is an important phenomenon to account for in causal graph generation. When causal pairs are naively connected into causal chains, transitive problems might occur by the contradictions between the threshold and scene factors in different instances of events \cite{xiong-etal-2022-reco}.
Generating causal links between events without accounting for the broader context in which they are embedded can lead to transitive incoherence.
In our setting, the events come from a specific context and the causal graphs are generated for the entire context.
The transitivity issues are likely more prevalent especially in settings where the event representations lack details---for example when using (verb, object) event representations~\cite{xiong-etal-2022-reco}  which lack connecting contexts. This in turn can lead to inconsistent predictions that do not respect transitivity\footnote{\citet{xiong-etal-2022-reco} fix this by adding back the contexts to the encoders used in their model.}.
In contrast, in our setting, the events are more grounded in the text and thus have richer connecting contexts which could reduce the likelihood of inconsistent relations. Moreover, we also take the context into account at each step of causal graph generation (our prompts nudge the models to consider the global graph structure) when making edge decisions. 





\section{Intrinsic Evaluation: Causal Graph Prediction}\label{sec:intrinsic-eval}
\begin{table*}
\centering
\begin{tabular}{@{}llcccc|cccc@{}}
\toprule
& \multirow{2}{*}{Model} & \multicolumn{4}{c}{Graph-level} & \multicolumn{4}{c}{Pair-level} \\ \cmidrule{3-6} \cmidrule{7-10} 
                      & & BAcc         & F1:C & F1:NC & Macro F1         & BAcc         &F1:C & F1:NC& Macro F1        \\ \midrule
\multirow{4}{*}{\rotatebox{90}{Llama}} & Direct &  63.08 &
53.42 & 69.35 & 61.39 & 63.96 & 53.70 & 70.50 & 62.10 \\
& Pairwise &  64.00	&41.39	&75.02	&58.20	&61.03 &42.09& 	70.64 &56.36\\
& Experts wo collab &  71.24 & 65.43 & 73.92 & 69.67 & 69.14 & 63.42 & 73.31 & 68.37
 \\
& Collab with experts &   73.69 & 73.31 & 71.67 & 72.49 & 72.07 & 70.87 & 73.17 & 72.02
\\
\midrule
\multirow{5}{*}{\rotatebox{90}{GPT-4o}} & Direct & 70.86 &  66.17 & 76.80 & 71.48 & 67.99 &  77.46 & 67.99 & 68.02 \\
& Pairwise & 73.93 &  62.99 & 82.37 & 72.68 & 74.08 &  67.29 & 81.67 & 74.48 \\
& Experts wo collab & 74.92 &  70.21 & 78.23 & 74.22 & 71.30 &  65.68 & 77.50 & 71.59 \\
& Collab with experts & \textbf{79.27}  & \textbf{75.62} & \textbf{82.80}& \textbf{79.21} & \textbf{77.12} &  \textbf{72.96} & \textbf{82.11} & \textbf{77.51} 
                       \\ \bottomrule
\end{tabular}
\caption{Performance of different approaches to causal graph generation using different LLMs. F1:C, F1:NC are the accuracies of identifying causal and non-causal links respectively. The proposed collaborative approach performs the best among all the settings, generating more accurate causal graphs.}
\label{tab:causal_graph_results}
\end{table*}

To show the effectiveness of our collaborative approach for producing accurate causal graphs, we first compare the quality of the generated graphs against those generated by a set of baselines. 

\subsection{Dataset}\label{crab-data}
The Causal Reasoning Assessment Benchmark (CRAB) \cite{romanou2023crab} is a dataset of $\sim$ 2.7k pairs for understanding causality of real-life events from news domain ({\color{black}for a justification on use of this dataset, please refer to Appendix \ref{crab-justification}).} 
%
%
CRAB contains causality scores for multiple event pairs in a document and therefore to use CRAB for our evaluation setup, we first group the pairs that are coming from the same article as context and then group them into causal and non-causal pairs according to their assigned scores (causal if above 50, otherwise non-causal). We also add the reverse of causal pairs to the list of non-causal pairs as causal links can not be bi-directional and that is a non-trivial condition that models should satisfy.

\subsection{Models}
{\color{black}Since the causal graph generation is a novel task, we had to design a set of baselines to showcase the performance of our collaborative approach.}
{\color{black}We compare the following approaches to generating causal graphs:
\textbf{(1) Direct:}
In this zero-shot setup similar to \cite{romanou2023crab}, we directly prompt the LLM to generate the full causal graph by generating all possible causal links.
\textbf{(2) Pairwise:}
This baseline, similar to the exhaustive pairwise approach used by \citet{long2023can}, goes over all possible event pairs (in both directions) and tries to identify whether the two events are causally linked. 
\textbf{(3) Experts wo collab:}
The goal of this setup is to show the importance of communication among experts towards the common goal of generating a comprehensive causal graph. In this setting, the experts each perform their assigned tasks but the results are aggregated without any communication among them.
This setup can be seen as a variation of chain of thought reasoning in which the model would first think of what are required as preconditions of causality and then generate the causal graph.
\textbf{(4) Collab wo experts:}
To show the effect of diverse expertise in generating causal graphs, we use a setup in which we only use causal experts (without any specific relation focus), all responsible for generating the causal graph.
\textbf{(5) Collab with experts:}
Our complete causal graph generation model consisting of $4$ relations experts collaborating through communicating with each other to generate a causal graph for a maximum of $3$ discussion rounds.
We have used GPT-4o \cite{openai2023gpt4} and Llama-70B-instruct \cite{llama3modelcard} as our base LLMs.}

\subsection{Results}
Our intrinsic evaluation focuses on analyzing different aspects of the collaborative approach for causal generation and how each component can help with the overall performance. We describe our findings in the following paragraphs.

\noindent\textbf{Causal link accuracy:}
The data points in CRAB \cite{romanou2023crab} are accompanied with gold labels (Causal and Non-causal as described in Section \ref{crab-data}).
We measured the performance of different approaches and reported balanced accuracy (BAcc) and Macro F1 ({\color{black}more details in Appendix \ref{app:eval-metrics}}) results for two settings. 
Since our ultimate goal is generating causal graphs that can later be used for downstream event reasoning tasks, we report the graph-level accuracy. For each causal graph associated with a scenario, we compute the BAcc and F1 scores (F1:C and F1:NC score of causal and non-causal links and Macro F1) and then report the average over all data points (graphs from CRAB) as shown in Table \ref{tab:causal_graph_results} under the graph-level numbers.
We also report the pair-level accuracy of identifying causal links to show how aiming for the causal graph can even lead to higher performance on pair level (Pair-level results in Table \ref{tab:causal_graph_results}).
As can be seen from Table \ref{tab:causal_graph_results}, the explicit help from relation experts and their collaboration results in better performance for both LLMs with higher accuracy and F1-scores. We also report an analysis on the effect of number of rounds on the performance in Appendix \ref{app:number_of_rounds}.

\begin{table*}
\centering
\begin{tabular}{@{}lcccc|cccc@{}}
\toprule
\multirow{2}{*}{Model} & \multicolumn{4}{c}{Graph-level} & \multicolumn{4}{c}{Pair-level} \\ \cmidrule{2-5} \cmidrule{6-9} 
                       & BAcc         & F1:C & F1:NC & Macro F1         & BAcc         &F1:C & F1:NC& Macro F1        \\ \midrule
Collab wo experts & 75.39 &  73.03 & 77.99 & 75.51 & 73.78 &  70.06 & 77.53 & 73.79 \\
Collab with experts & \textbf{79.27}  & \textbf{75.62} & \textbf{82.80}& \textbf{79.21} & \textbf{77.12} &  \textbf{72.96} & \textbf{82.11} & \textbf{77.51} \\
\midrule
Collab wo temporal & 77.51 & 74.75 & 80.77 & 77.72 & 74.96 & 70.50& 79.99 & 75.25 \\
Collab wo discourse & 78.32 & 74.67 & 81.92 & 78.29 & 74.72 & 68.89& 80.28 & 75.09 \\
Collab wo precondition & 77.48 & 73.14 & 81.39&77.26 & 74.08 &  69.42 & 79.33 & 74.37 \\
Collab wo common sense & 78.88 &  74.68 & 82.81 & 78.85 & 74.33 &  69.36 & 80.05 & 74.71 

                       \\ \bottomrule
\end{tabular}
\caption{Performance of different experts on causal graph generation using GPT-4o as the base LLM.}
\label{tab:causal_graph_ablations}
\end{table*}
{\color{black}\noindent\textbf{Experts and collaboration effect:}
The main components of our approach are the diverse relation experts and collaboration among them. Collaboration allows experts to communicate with each other to come to a shared causal representation of the scenario. Not giving the experts a chance to talk to each other, falls short on fully exploiting their potential. We compare two settings with and without the collaboration among experts (experts wo collab and collab w experts in Table \ref{tab:causal_graph_results}) and show how the performance would be degraded without proper communication.
We have identified a group of experts whom their expertise can help with the causal graph identification. 
As can be seen from first section of Table \ref{tab:causal_graph_ablations}, causal graph generation without the presence of experts is lacking which highlights the importance of such relations experts. But how important is each expert? We conduct ablations of our approach each time by removing one of the experts and observed drops in performance as a common trend for all experts (second section of Table \ref{tab:causal_graph_ablations}). 
\begin{table*}
\small
\begin{tabular}{@{}p{1.8cm}|p{1.9cm}|p{1.9cm}|p{1.9cm}|p{0.5cm}|p{1cm}|p{1.8cm}|p{1.5cm}@{}}
\toprule
Agent & Initial overlap \newline with gold pairs & Final overlap \newline with gold pairs & Contribution to \newline Identified pairs & Total flips & Incorrect flips & Additions \newline from experts & Conflicting agents \\ 
\midrule
Temporal & 13\%\ & 33\%& 64\% & 24\% & 0\%&17\%& C-P-D
\\
Discourse & 17\% &24\%& 64\% & 9\% & 0\% & 39\%& T-C-P
\\
Precondition & 17\% &22\%& 46\% & 22\%& 67\%& 38\%& T-D-C
\\
Commonsense & 22\% &26\%& 57\% & 0\% &0\%& 19\%& T-D-P
                           \\ \bottomrule
\end{tabular}
\caption{Analysis of experts debates. We report how accurate each expert is in the beginning and in the end and how much it would contribute to the final list of causal edges. The flips and additions (to measure persuasiveness by other experts) are also shown. The last column represents the conflicting experts for each individual expert (from highest to lowest). \textit{T}, \textit{D}, \textit{P} and \textit{C} stand for temporal, discourse, precondition and commonsense respectively.}
\label{tab:agents_analysis}
\end{table*}
We also conducted another set of ablations to see how individual experts perform on their own. We report the results in Table \ref{tab:causal_graph_experts_ablations} in Appendix \ref{app:individual-experts}.
While individual experts perform better than direct baseline, it is only through collaboration among experts that their full potential is exploited and the best performance can be achieved.

\noindent\textbf{Collaboration trajectories analysis:}
We looked at the trajectories of different debate sessions (selected 10\%) and analyzed how different agents behaved during debates. 
Below are some of the findings that are reported in Table \ref{tab:agents_analysis}. As described earlier, experts first generate their own responses and then engage in a debate.
We first measure how accurate each expert's initial response is (second column) with temporal expert being the weakest expert. 
The collaboration is aimed at improving the performance of experts so as shown in the third column of Table \ref{tab:agents_analysis}, all experts perform better once the debate is over with temporal expert having the highest gain and becoming the most accurate expert. 
Another aspect analyzed is how much each expert contributes to the final list of causal graphs. We measured the overlap between each expert's initial set and the final set of causal links in the third column. While the precondition expert starts with a reasonable performance, due to its high number of incorrect flips (5th and 6th columns), it would not contribute as much as the other experts.
Also, discourse and precondition experts have more additions (more edges identified compared to the initial round), showing they are more influenced by other experts. 
Finally, we measured conflict: how much different experts disagree on their decisions. For each expert, we reported a conflict ordering in which experts with high conflict rank higher.  Precondition and commonsense experts have lower conflict on their identified pairs whereas they have the highest disagreement with the temporal expert.}

We also conducted the cost analysis of different approaches to causal graph generation and report the results in Appendix \ref{app:cost-analysis} and studied the effect of the number of experts in our collaborative approach in Appendix \ref{app:experts-number-analysis}.
\section{Extrinsic Evaluation: Event Reasoning with Causal Graphs}
\label{sec:extrinsic-eval}

We show the effectiveness of using a universal causal graph structure on a set of downstream event reasoning applications.
We first introduce a new event reasoning task, Explainable Event Likelihood prediction, \task, to address the lack of interpretability of existing event modeling approaches and then show how using a causal graph as the global backbone can help with this new task on three criteria: causality, informativeness, and coherence.
Aside from this novel reasoning task, we also assess how well event likelihood prediction with causal graphs can perform on downstream applications: event forecasting and next event prediction and show that \task with causal graph (with no task-specific fine-tuning) is competitive to the SOTA methods.

\subsection{Causally Explainable Event Likelihood Prediction}
{\color{black}
Traditionally, event likelihood is modeled as a next event prediction task in which a system generates events that are likely to co-occur within a given event context. This framing, however, doesn't provide the explanation for the likelihood prediction itself. 
%
Without clear \emph{causal} explanations, it is difficult to understand what logical connections to the observed events support a predicted likelihood. For example, in a flood scenario, human schematic knowledge might list ``people leave affected areas'' as a plausible event, but its actual likelihood depends on many factors, such as the severity and damage of the flood. Minor flood does not include this event, but severe does. A model's likelihood prediction for an event must be determined and justified by causal relations with context events.}

To account for this, we introduce \task\ where the input is a body of text ($D$) that contains a collection of observed events $E = <e_1, \cdots, e_n>$, and a queried event of interest $e_q$ that is not directly mentioned in the text. The system is required to predict whether $e_q$ is likely or unlikely, and also provide a subset of events from the input context that can be interpreted as causal explanations for the likelihood of the event. The system is expected to present the causal explanations as a chain of events.
%
Our approach to addressing these issues is to effectively structure the model's reasoning to systematically reflect upon the global causal structure of the events. To this end, we use events causal graph to predict whether a query event is likely.
A query event is deemed likely if it can be inserted into the graph.
We thus ask the model to place the query event in the causal graph, by considering which events are causes and which are effects. 
We operationalize this idea using a process as shown in Figure \ref{fig:overview} and refer this Causal Graph-based Event Likelihood prediction as \method. The details of all steps are further described in the Appendix \ref{appendix:eel}.



\subsection{Experimental Setting}\label{eel_exp}
We curated a small test set (520 \textit{(news, query event)} pairs) using the Annotated NYT corpus \cite{sandhaus2008new}.
More details on the curation process available in the Appendix \ref{append:data}.
We also report the results using three metrics: \textbf{\textit{Causality}}, \textbf{\textit{Informativeness}}, and
\textbf{\textit{Coherence}} (More details of this novel 
evaluation setup is described in the Appendix \ref{app:criteria}).

\begin{table*}
\centering
\small
\begin{tabular}{l|lll|lll|lll}
\multirow{2}{*}{}             & \multicolumn{3}{c|}{\textbf{Causality}} & \multicolumn{3}{c|}{\textbf{Informativeness}} & \multicolumn{3}{c}{\textbf{Coherence}} \\
                              & \multicolumn{1}{c}{\small{Baseline}} & \multicolumn{1}{c}{\small{\method}} & \multicolumn{1}{c|}{\small{Tie}} & \multicolumn{1}{c}{\small{Baseline}} & \multicolumn{1}{c}{\small{\method}} & \multicolumn{1}{c|}{\small{Tie}} & \multicolumn{1}{c}{\small{Baseline}} & \multicolumn{1}{c}{\small{\method}} & \multicolumn{1}{c}{\small{Tie}} \\
\hline
One-shot             & 21.1 & 41.6 & 37.3 & 31.9 & 48.4 & 19.6 & 30.0 & 37.0 & 33.0\\
One-shot w events    & 18.1 & 32.5 & 49.4 & 22.9 & 52.9 & 24.2 & 14.4 & 47.3 & 38.3 \\
One-shot w Reflexion & 18.8 & 39.3 & 41.9 & 30.1 & 49.9 & 20.1 & 29.9 & 33.8 & 36.3                         
\end{tabular}
\caption{Pairwise comparison of 3 baseline systems vs our causal chain approach (\method) on generating event likelihood explanations. Each row compares the performance of one baseline to \method\ on $3$ evaluation metrics. Each column's number is the percentage of times that model was judged best. `Tie' means they scored equally.}
\label{tab:main-table}
\end{table*}
\begin{table}
\centering
\small
\begin{tabular}{@{}lc@{}}
\toprule
\multicolumn{1}{c}{System}    & Accuracy \\ \midrule
GPT-4 baseline                & 51.3                 \\
One-shot baseline                & 50.0                 \\ 
\method                  & 62.7                 \\
BERT base + MDS$^*$  & 63.1                 \\
BERT large + MDS$^*$ & 67.4                 \\
Human performance             & 74.6                 \\ \bottomrule
\end{tabular}
\caption{Forecasting accuracy of different systems. Models with $^*$ are from \cite{jin2021forecastqa} that have the most similar setting as what we used for the other systems. Systems were tested on 25\% of the forecastQA data to reduce cost associated with using GPT-4.} 
\label{tab:forecasting}
\end{table}


\subsubsection{Results}

We compare \method with 3 baselines; one-shot causal chain, one-shot causal chain with events and one-shot causal chain with reflection \cite{shinn2023reflexion} (which are described in detail in Appendix \ref{app:methods}) and report the results in Table~\ref{tab:main-table}. 
The evaluation is a pairwise comparison of two systems on each evaluation criterion to see which system is better.
Each row represents a baseline system against our causal graph-based approach (\method), and for each evaluation dimension, we report the percentage of times each system generated a better chain. 
For each instance in the test, the winner can be baseline, CGEL, or Tie. 
As seen in Table~\ref{tab:main-table}, the graph-based causal chain approach wins the most in all settings over all baselines with its strongest performance on Informativeness.
%
The one-shot baseline setting which uses the same set of events extracted for causal graph generation performs the best among baselines in terms of causality. Providing events to the system might make it more focused towards generating the correct causal links between event. 
%
The one-shot baseline with reflection, in which we ask the system to reflect on its generated causal chains and update them in an additional step, performs slightly better than the other baseline settings. 
%
More results and analysis on comparing GPT-4 evaluation versus human evaluations, the effect of different heuristics selection strategies on the performance and an error analysis is presented in the Appendix \ref{more_results}.

\subsection{Event Forecasting}
Event forecasting \cite{jin2021forecastqa} is the challenging task of predicting future events to better help with planning.
While not designed necessarily with forecasting in mind, our causal graph-based approach to event likelihood prediction can be applied to the forecasting task; an event will be considered likely to occur in future if it can be placed in the causal graph as a leaf node.
{\color{black}We processed ForecastQA dataset \cite{jin2021forecastqa} as described in Appendix \ref{app:forecast:data}.}
%
We compared the one-shot baseline chain and \method for each question and checked whether the event in question can be placed as a leaf node in the graph. For the \textit{yes} answer questions, the model is deemed correct if it places the event as a leaf node, and for \textit{no} answer questions it is correct if it doesn't place it in the graph.
We measure the accuracy of the causal approaches and report the results in Table \ref{tab:forecasting}. We also added another baseline in which we directly ask GPT-4 to answer the question given the context. We further report the numbers for fine-tuned models used in \citet{jin2021forecastqa} with similar settings to our approach (using the text summarizer to shorten the context). 
As seen from Table \ref{tab:forecasting}, our graph-based causal approach again significantly outperforms the GPT-4 baseline.
Further, it achieves competitive results when compared to models fine-tuned on the dataset.
In addition, graph-based approach provides \emph{explanations} as to why the event is likely to occur in future, which is not the case for the fine-tuned models (examples in Table \ref{tab:forecast-examples}).
\subsection{Next Event Prediction}
We also compare the accuracy for next event prediction framed as a multiple-choice cloze task in which each system has to correctly predict the gold output from a fixed set of events \cite{koupaee-etal-2023-modeling}. 
Using \method, the system makes the right decision if and only if it places the correct event in the causal graph. The prediction is incorrect if it also places any of the incorrect events in the causal graph.
We used the same checkpoints and the same data from~\citet{koupaee-etal-2023-modeling} and computed the accuracy on a subset of the original test set (10\% to reduce the costs associated with using GPT-4).
The results are shown in Table \ref{tab:nc}. There are two versions of the one-shot and \method approach; one with only events and one with the context the events are coming from (please note that in this version, the context excludes sentences that contain the query event).
As can be seen from Table \ref{tab:nc}, \method is capable of predicting co-occurring events better than the one-shot baseline showing the effectiveness of causal graphs. Moreover, our approach is competitive to all event language models that have been 
specifically trained to do this task. 
\section{Related Work}


%

\noindent\textbf{Event Relations:} 
There has been a body of work on extracting different types of semantic relations between events, including but not limited to co-occurrence \cite{modi2016event,pichotta2016learning}, temporal \cite{pustejovsky2003timebank,jin2023toward}, causal \cite{pustejovsky2003timebank}, conditional \cite{kwon2020modeling}, counterfactual, etc \cite{sap2019atomic,han2021ester}.
Most of the prior works on events relations only take the local pairwise connections into account without considering the global picture. 
Even when considering the complex relations between events, the events are usually connected through discourse or temporal relations \cite{li2021future}.
However, it has been shown how a global causal picture can help with story understanding \cite{sun2023event}.

Our approach is aimed at identifying causal relations between instantiated events while taking their context into account.
This differs from statistical or rule-based approaches \cite{gordon2011commonsense, luo2016commonsense} which focus on surface-level associations to measure causality between pairwise events. In our approach, experts are responsible for taking the semantics of events into account through various relations and then decide whether the link is causal.

\begin{table}
\centering
\small
\begin{tabular}{@{}l|c@{}}
\toprule
\multicolumn{1}{c}{System}    & Accuracy \\ \midrule
ELM                & 46.0               \\
EGELM                & 50.0                 \\
QGELM & 46.0 \\
\midrule
One-shot baseline                 & 13.0                 \\ 
\method                  & 55.0                 \\
\midrule
One-shot baseline with context              & 38.0                 \\ 

\method with context              & 61.0                 \\ \bottomrule
\end{tabular}
\caption{NC accuracy of different systems.} 
\label{tab:nc}
\end{table}

\noindent\textbf{Causal Graph for Event Reasoning:} 
Event language modeling has been used as a means to induce event schemas, as a connected sets of events and their actors. More specifically, event language models aim at predicting the next events given a context of events \cite{manshadi2008learning, pichotta2016learning, modi2016event, weber2018hierarchical,rezaee2021event} that can be used to reason about real-life events.
However, to capture more complex relations within a scenario, graph-based approaches \cite{li2021future,li-etal-2020-connecting} are used to model multi-dimensional relations between the events and the entities .
Event graphs have been applied for event schema induction, to better characterize temporal and multi-hop argument relations \cite{li-etal-2020-connecting, li-etal-2021-future}, and capture the global dependencies between events \cite{jin-etal-2022-event}. In contrast, our work constructs event graphs with causal connections.  Event Scripts are not a simple list of events but rather linked causal chains~\cite{pearl1995causal}. However, the event likelihood prediction based solely on the correlational measures such as $p(e_2|e_1)$ does not account for causal relevance~\cite{weber-etal-2020-causal}, and can lead to spurious event predictions~\cite{ge2016discovering}. 
To address this, \citet{weber-etal-2020-causal} consider the causal \emph{relevance} in event predictions, but they do not explicitly model the causal \emph{structure} between events. 
Compared to unstructured causal event pairs \cite{luo2016commonsense}, causal event graphs explicitly consider the structure of causal relationships, capture interdependence among causalities \cite{wang-etal-2022-care} for emotional label inference, and make the causal reasoning explainable for COPA entailment tasks \cite{du-etal-2021-excar}. 
Given a causal graph, LLMs can help with generating narratives in form of text descriptions \cite{phatak2024narrating}. Moreover, 
LLMs can be used directly to extract causal relations and even generate causal graphs by looping over all possible pairs of events (i.e. nodes in the output graph) and assessing whether they share a causal link \cite{long2023can}. However, we show that this enumerative approach to causal graph generation is less effective compared to our collaborative approach which considers diverse reasoning strategies.

\noindent\textbf{LLM-based Agents:}
There is a large body of work on multi-LLM setups where the main difference is the task distribution and the collaboration mechanism.
In the first group, a task is broken into multiple sub-tasks and each LLM is asked to perform its assigned task that can help with realizing the overall goal \cite{hong2023metagpt,mandi2024roco,qian2024chatdev}. 
In the second category, all LLMs are assigned the same task with either the same or different roles and personas \cite{liang2023encouraging, du2023improving, chan2023chateval,lan2024stance}. In this setup, the agents might also communicate with each other to achieve a final goal. The conversation among agents would force agents to dig deeper and come up with more sound arguments and also give them a chance to modify their responses based on other agents' arguments.
Depending on the task, the final response would be either the majority vote or the final ruling of a judge LLM.


\section{Conclusion}
{\color{black}
To address the need for a causal structure for event reasoning,
we introduce a collaborative multi-turn approach for causal graph generation. This approach induces diverse reasoning approaches that correspond to specific types of semantic relations leading to better causal graphs overall.
We also show that the causal graphs thus generated have better utility in downstream event reasoning tasks such as our novel explainable event likelihood prediction, next event prediction, and event forecasting.}

\section*{Limitations}
Our approach depends on the performance of LLMs (GPT-4 and Llama) and their internal understanding of causality which may be different from how humans would perceive causality in some situations.
Furthermore, our evaluation on event likelihood prediction is done using GPT-4 and human annotators. Although GPT-4 has been used as an evaluator and we have shown that the results do not change much using human annotators vs GPT-4, that automatic evaluation may still exhibit bias towards GPT-4. Also, our work doesn't capture a notion of causality strength. Recent work has shown that humans can provided graded judgments of causality \cite{romanou2023crab}. Modeling graded judgments can provide a richer representation for reasoning and even allow similar graded judgments of the event likelihood as well.  

\section*{Acknowledgments}
We thank the anonymous reviewers for their insightful
feedback and suggestions.
This material is based on research that is in part supported by DARPA for the SciFy program under agreement number HR00112520301. The U.S. Government is authorized to reproduce and distribute reprints for Governmental purposes notwithstanding any copyright notation thereon. 
%
%
The views expressed in this paper are those of the author(s) and do not reflect the official policy or position of DARPA, the U.S. Naval Academy, Department of the Navy, the Department of Defense, or the U.S. Government.

\bibliography{custom}
\newpage
\clearpage
\appendix
\section{Collaborative Causal Graph Generation}

{\color{black}
\subsection{Dataset}\label{crab-justification}
We used CRAB \cite{romanou2023crab} which consists of $\sim$ 2.7k pairs annotated with causality scores from 0-100.
The main thing to note first is that there is a lack of datasets with annotated causal graphs from context in any domain. 
As far as the authors know, the existing causal datasets used in the NLP community (e.g. COPA \cite{roemmele2011choice}, eCARE \cite{du2205care} ) except for the CRAB dataset used in our evaluations, are limited to pairwise annotation and thus don’t capture the broad picture of causality over the full context as this is what we aim to do here. 

News events have long been an important application area for event reasoning since they often cover a range of domains within themselves (entertainment, politics, sports, etc.), hence offer a diverse testbed of events for our experiments.}

\subsection{Evaluation Criteria}\label{app:eval-metrics}
\paragraph{BAcc}
The data points in this dataset are accompanied with gold labels (Causal and Non-causal). We then use the gold labels and measure F1 (described in Appendix A.1) by comparing the predicted labels (outputs on whether two events have a causal link or not, from different settings we have tried) and the gold labels. 
{\color{black}
The balanced accuracy is the average between the sensitivity (true positive rate) and the specificity (true negative rate), which measures the average accuracy obtained from both the causal and non-causal classes. 

\begin{equation}
\text{BAcc} = (\text{sensitivity} + \text{specificity})/2
\end{equation}

\noindent{Where sensitivity or TPR = TP / (TP + FN) and specificity or TNR = TN / (TN + FP).}

\noindent{\textbf{TP} = The number of times the predicted label is causal and the human label from the dataset is also causal}

\noindent{\textbf{TN} = The number of times the predicted label is non-causal and the human label from the dataset is also non-causal}

\noindent{\textbf{FP} = The number of times the predicted label is causal whereas the human label from the dataset is non-causal}

\noindent{\textbf{FN} = The number of times the predicted label is non-causal whereas the human label from the dataset is causal}}

\paragraph{Macro F1 score}
We also report the Macro F1 scores by computing F1 scores for both the causal and non-causal classes. 

\subsection{Prompts}\label{causal-graph-prompts}
The prompts that we have used for our approach are presented in following tables: Temporal expert in Table \ref{tab:temporal_prompt}, discourse expert in Table \ref{tab:discourse_prompt}, precondition expert in \ref{tab:precondition_prompt}, commonsense expert in \ref{tab:commonsense_prompt} and finally the causal judge in \ref{tab:judge_prompt}.
\begin{table*}
\centering
\small
\begin{tabular}{@{}p{14cm}@{}}
\toprule
Given the following document:
\newline
[NEWS]
\newline
\newline
Events:
\newline
[EVENTS]
\newline
\newline
Identify all causal pairs with first event causing the second one. You are responsible for looking into the temporality of the given events and evaluate the causality from the temporal perspective. There are also other evaluator experts assigned the same task as you.
\newline
\newline
You are evaluating this pair along with 3 other experts, each responsible to check the pair for a specific aspect that can help with identifying causality.
\newline
1. Temporal expert (you), responsible for taking into account the temporal relations as prerequisites for a causal relation.
\newline
2. Precondition expert, responsible for checking how one event would affect the other event, mainly if removing one of the events would also lead to making the other one irrelevant.
\newline
3. Commonsense expert, responsible for identifying the missing commonsense bits that can help with identifying whether there is a causal link or not.
\newline
4. Discourse expert, responsible for identifying whether events are sharing entities that can lead to identifying causal links.
\newline
\newline
You can see your previous responses as well as other experts responses. You can see the list of all identified pairs by all experts. Continue the discussion with other evaluator experts, talk to them and state why you agree/disagree with each other bringing as many arguments as you can while remembering your main task which is considering the temporal aspect.
\newline
\newline
Discussion history:
\newline
[MESSAGES]
\newline
\newline
Causal pairs:
\newline
[CAUSAL PAIRS]
\newline
\newline

Find any new causal relation for the given events. Please make sure that the newly added pairs do not violate the existing ones. Note that it is not possible to have both a causes b and b causes a. Also note that if a causes b and b causes c, then a causes c as well. Provide the ids of each pair of events with causal relation in <pair></pair> where the first event is the cause and the second is the effect and then place all pairs in separate lines in <pairs></pairs> tags.
    \newline
    \newline
    Example of output format:
    \newline
    <pairs>
    \newline
    <pair>1,2</pair>
    \newline
    <pair>3,4</pair>
    \newline
    </pairs>
    \\ 
 \bottomrule
\end{tabular}
\caption{Prompt used for temporal expert.}
\label{tab:temporal_prompt}
\end{table*}
\begin{table*}
\centering
\small
\begin{tabular}{@{}p{14cm}@{}}
\toprule
Given the following document:
\newline
[NEWS]
\newline
\newline
Events:
\newline
[EVENTS]
\newline
\newline
Identify all causal pairs with first event causing the second one. You are responsible for looking into common entities that are shared among the given events and evaluate the causality from that perspective. Events with causal relation can potentially be sharing some entities. There are also other evaluator experts assigned the same task as you.
\newline
\newline
You are evaluating this pair along with 3 other experts, each responsible to check the pair for a specific aspect that can help with identifying causality.
\newline
1. Discourse expert (you), responsible for identifying whether events are sharing entities that can lead to identifying causal links.
\newline
2. Temporal expert, responsible for taking into account the temporal relations as prerequisites for a causal relation.
\newline
3. Commonsense expert, responsible for identifying the missing commonsense bits that can help with identifying whether there is a causal link or not.
\newline
4. Precondition expert, responsible for checking how one event would affect the other event, mainly if removing one of the events would also lead to making the other one irrelevant.
\newline
\newline
You can see your previous responses as well as other experts responses. You can see the list of all identified pairs by all experts. Continue the discussion with other evaluator experts, talk to them and state why you agree/disagree with each other bringing as many arguments as you can while remembering your main task which is considering the entity sharing aspect of the given events.
\newline
\newline
Discussion history:
\newline
[MESSAGES]
\newline
\newline
Causal pairs:
\newline
[CAUSAL PAIRS]
\newline
\newline

Find any new causal relation for the given events. Please make sure that the newly added pairs do not violate the existing ones. Note that it is not possible to have both a causes b and b causes a. Also note that if a causes b and b causes c, then a causes c as well. Provide the ids of each pair of events with causal relation in <pair></pair> where the first event is the cause and the second is the effect and then place all pairs in separate lines in <pairs></pairs> tags.
    \newline
    \newline
    Example of output format:
    \newline
    <pairs>
    \newline
    <pair>1,2</pair>
    \newline
    <pair>3,4</pair>
    \newline
    </pairs>
    \\ 
 \bottomrule
\end{tabular}
\caption{Prompt used for discourse expert.}
\label{tab:discourse_prompt}
\end{table*}
\begin{table*}
\centering
\small
\begin{tabular}{@{}p{14cm}@{}}
\toprule
Given the following document:
\newline
[NEWS]
\newline
\newline
Events:
\newline
[EVENTS]
\newline
\newline
Identify all causal pairs with first event causing the second one. You are responsible for looking into the preconditional dependence of the given events and evaluate the causality from the dependence perspective. Events are dependent if removing one of them (assuming it did not happen) leads to making the other event irrelevant. There are also other evaluator experts assigned the same task as you.
\newline
\newline
You are evaluating this pair along with 3 other experts, each responsible to check the pair for a specific aspect that can help with identifying causality.
\newline
1. Precondition expert (you), responsible for checking how one event would affect the other event, mainly if removing one of the events would also lead to making the other one irrelevant.
\newline
2. Temporal expert, responsible for taking into account the temporal relations as prerequisites for a causal relation.
\newline
3. Commonsense expert, responsible for identifying the missing commonsense bits that can help with identifying whether there is a causal link or not.
\newline
4. Discourse expert, responsible for identifying whether events are sharing entities that can lead to identifying causal links.
\newline
\newline
You can see your previous responses as well as other experts responses. You can see the list of all identified pairs by all experts. Continue the discussion with other evaluator experts, talk to them and state why you agree/disagree with each other bringing as many arguments as you can while remembering your main task which is considering the dependence of the given events.
\newline
\newline
Discussion history:
\newline
[MESSAGES]
\newline
\newline
Causal pairs:
\newline
[CAUSAL PAIRS]
\newline
\newline

Find any new causal relation for the given events. Please make sure that the newly added pairs do not violate the existing ones. Note that it is not possible to have both a causes b and b causes a. Also note that if a causes b and b causes c, then a causes c as well. Provide the ids of each pair of events with causal relation in <pair></pair> where the first event is the cause and the second is the effect and then place all pairs in separate lines in <pairs></pairs> tags.
    \newline
    \newline
    Example of output format:
    \newline
    <pairs>
    \newline
    <pair>1,2</pair>
    \newline
    <pair>3,4</pair>
    \newline
    </pairs>
    \\ 
 \bottomrule
\end{tabular}
\caption{Prompt used for precondition expert.}
\label{tab:precondition_prompt}
\end{table*}
\begin{table*}
\centering
\small
\begin{tabular}{@{}p{14cm}@{}}
\toprule
Given the following document:
\newline
[NEWS]
\newline
\newline
Events:
\newline
[EVENTS]
\newline
\newline
Identify all causal pairs with first event causing the second one. You are responsible for looking into identifying the commonsense that is relevant to the given events and use them to identify whether there is a causal link or not. The missing commonsense can be some intermediate events that can help with identifying the link between the given pair of events. There are also other evaluator experts assigned the same task as you.
\newline
\newline
You are evaluating this pair along with 3 other experts, each responsible to check the pair for a specific aspect that can help with identifying causality.
\newline
1. Commonsense expert (you), responsible for identifying the missing commonsense bits that can help with identifying whether there is a causal link or not.
\newline
2. Precondition expert, responsible for checking how one event would affect the other event, mainly if removing one of the events would also lead to making the other one irrelevant.
\newline
3. Temporal expert, responsible for taking into account the temporal relations as prerequisites for a causal relation.
\newline
4. Discourse expert, responsible for identifying whether events are sharing entities that can lead to identifying causal links.
\newline
\newline
You can see your previous responses as well as other experts responses. You can see the list of all identified pairs by all experts. Continue the discussion with other evaluator experts, talk to them and state why you agree/disagree with each other bringing as many arguments as you can while remembering your main task which is using commonsense to identify the causal link.
\newline
\newline
Discussion history:
\newline
[MESSAGES]
\newline
\newline
Causal pairs:
\newline
[CAUSAL PAIRS]
\newline
\newline

Find any new causal relation for the given events. Please make sure that the newly added pairs do not violate the existing ones. Note that it is not possible to have both a causes b and b causes a. Also note that if a causes b and b causes c, then a causes c as well. Provide the ids of each pair of events with causal relation in <pair></pair> where the first event is the cause and the second is the effect and then place all pairs in separate lines in <pairs></pairs> tags.
    \newline
    \newline
    Example of output format:
    \newline
    <pairs>
    \newline
    <pair>1,2</pair>
    \newline
    <pair>3,4</pair>
    \newline
    </pairs>
    \\ 
 \bottomrule
\end{tabular}
\caption{Prompt used for commonsense expert.}
\label{tab:commonsense_prompt}
\end{table*}
\begin{table*}
\centering
\small
\begin{tabular}{@{}p{14cm}@{}}
\toprule
Given the following document:
\newline
[NEWS]
\newline
\newline
Events:
\newline
[EVENTS]
\newline
\newline
Causal pairs are generated by 4 experts, each responsible to check for a specific aspect that can help with identifying causality.
\newline
1. Temporal expert, responsible for taking into account the temporal relations as prerequisites for a causal relation.
\newline
2. Precondition expert, responsible for checking how one event would affect the other event, mainly if removing one of the events would also lead to making the other one irrelevant.
\newline
3. Commonsense expert, responsible for identifying the missing commonsense bits that can help with identifying whether there is a causal link or not.
\newline
4. Discourse expert, responsible for identifying whether events are sharing entities that can lead to identifying causal links.
\newline
\newline
You can see the discussion history and the list of all identified pairs by all experts. 
\newline
\newline
Discussion history:
\newline
[MESSAGES]
\newline
\newline
Causal pairs:
\newline
[CAUSAL PAIRS]
\newline
\newline
Finalize the list of causal relations for the given events and be comprehensive about that. Note that if 1 causes 2 and 2 causes 3, then 1 causes 3 as well, therefore add all such relations as well. Provide the ids of each pair of events with causal relation in <pair></pair> where the first event is the cause and the second is the effect and then place all pairs in separate lines in <pairs></pairs> tags.
    \newline
    \newline
    Example of output format:
    \newline
    <pairs>
    \newline
    <pair>1,2</pair>
    \newline
    <pair>3,4</pair>
    \newline
    </pairs>
    \\ 
 \bottomrule
\end{tabular}
\caption{Prompt used for causal judge.}
\label{tab:judge_prompt}
\end{table*}
\subsection{More Analysis on Causal Graphs}\label{app:number_of_rounds}
\subsubsection{Collaboration Round Effect}
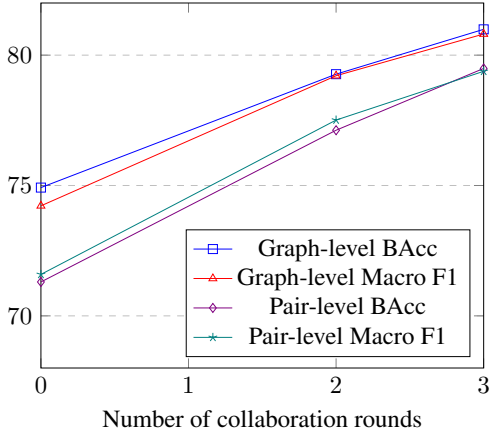
\begin{figure}
    \centering    
\begin{tikzpicture}[scale=0.85]
\begin{axis}[
    bar width=0.7cm,
    xlabel={Number of collaboration rounds},
    scaled y ticks = false,
    xmin=0, xmax=3,
    ymin=68, ymax=82,
    xtick={0,1,2,3},
    legend pos=south east,
    ymajorgrids=true,
    grid style=dashed,
]

\addplot[
    color=blue,
    mark=square,
    ]
    coordinates {
    (0,74.92)(2,79.27)(3,80.99)
    };

\addplot[
    color=red,
    mark=triangle,
    ]
    coordinates {
    (0,74.22)(2,79.21)(3,80.81)
    };

    \addplot[
    color=violet,
    mark=diamond,
    ]
    coordinates {
    (0,71.30)(2,77.12)(3,79.49)
    };

\addplot[
    color=teal,
    mark=star,
    ]
    coordinates {
    (0,71.59)(2,77.51)(3,79.38)
    };
    \legend{Graph-level BAcc, Graph-level Macro F1, Pair-level BAcc, Pair-level Macro F1}  
\end{axis}
\end{tikzpicture}
\caption{The effect of collaboration rounds on performance.}
    \label{fig:rounds_effect}
\end{figure}

To show how different number of collaboration rounds can affect the overall performance, we measured BAcc and Macro F1 with respect to the number of communications among experts and plotted the results in Figure \ref{fig:rounds_effect}. Both metrics tend to increase by having more communication rounds however, this increase is not linear and the plot gradually starts to plateau. 

{\color{black}\subsubsection{Individual Experts Performance}\label{app:individual-experts}
To further evaluate the role of each expert in making causal judgments we conducted another ablation study.
Each time, the LLM is tasked to generate the causal graph with respect to only one of the expertise described in Section \ref{sec:relations-epxerts}. 
\begin{table*}
\centering
\small
\begin{tabular}{@{}p{14cm}@{}}
\toprule
Given the following document:
\newline
[NEWS]
\newline
\newline
Events:
\newline
[EVENTS]
\newline
\newline
Identify all causal pairs with first event causing the second one. You are responsible for looking into the temporality of the given events and evaluate the causality from the temporal perspective.
\newline
\newline
Place your reasoning between <argument></argument> tags. 
Find any new causal relation for the given events. Please make sure that the newly added pairs do not violate the existing ones. Note that it is not possible to have both a causes b and b causes a. Also note that if a causes b and b causes c, then a causes c as well. Provide the ids of each pair of events with causal relation in <pair></pair> where the first event is the cause and the second is the effect and then place all pairs in separate lines in <pairs></pairs> tags.
    \newline
    \newline
    Example of output format:
    \newline
    <pairs>
    \newline
    <pair>1,2</pair>
    \newline
    <pair>3,4</pair>
    \newline
    </pairs>
    \\ 
 \bottomrule
\end{tabular}
\caption{Prompt used for agent with temporal expertise.}
\label{tab:temporal_prompt_single}
\end{table*}
\begin{table*}
\centering
\small
\begin{tabular}{@{}p{14cm}@{}}
\toprule
Given the following document:
\newline
[NEWS]
\newline
\newline
Events:
\newline
[EVENTS]
\newline
\newline
Identify all causal pairs with first event causing the second one. You are responsible for looking into common entities that are shared among the given events and evaluate the causality from that perspective. Events with causal relation can potentially be sharing some entities.
\newline
\newline
Place your reasoning between <argument></argument> tags. 
Find any new causal relation for the given events. Please make sure that the newly added pairs do not violate the existing ones. Note that it is not possible to have both a causes b and b causes a. Also note that if a causes b and b causes c, then a causes c as well. Provide the ids of each pair of events with causal relation in <pair></pair> where the first event is the cause and the second is the effect and then place all pairs in separate lines in <pairs></pairs> tags.
    \newline
    \newline
    Example of output format:
    \newline
    <pairs>
    \newline
    <pair>1,2</pair>
    \newline
    <pair>3,4</pair>
    \newline
    </pairs>
    \\ 
 \bottomrule
\end{tabular}
\caption{Prompt used for agent with discourse expertise.}
\label{tab:discourse_prompt_single}
\end{table*}
\begin{table*}
\centering
\small
\begin{tabular}{@{}p{14cm}@{}}
\toprule
Given the following document:
\newline
[NEWS]
\newline
\newline
Events:
\newline
[EVENTS]
\newline
\newline
You are responsible for looking into the preconditional dependence of the given events and evaluate the causality from the dependence perspective. Events are dependent if removing one of them (assuming it did not happen) leads to making the other event irrelevant.
\newline
\newline
Place your reasoning between <argument></argument> tags. 
Find any new causal relation for the given events. Please make sure that the newly added pairs do not violate the existing ones. Note that it is not possible to have both a causes b and b causes a. Also note that if a causes b and b causes c, then a causes c as well. Provide the ids of each pair of events with causal relation in <pair></pair> where the first event is the cause and the second is the effect and then place all pairs in separate lines in <pairs></pairs> tags.
    \newline
    \newline
    Example of output format:
    \newline
    <pairs>
    \newline
    <pair>1,2</pair>
    \newline
    <pair>3,4</pair>
    \newline
    </pairs>
    \\ 
 \bottomrule
\end{tabular}
\caption{Prompt used for agent with precondition expertise.}
\label{tab:precondition_prompt_single}
\end{table*}
\begin{table*}
\color{black}
\centering
\small
\begin{tabular}{@{}p{14cm}@{}}
\toprule
Given the following document:
\newline
[NEWS]
\newline
\newline
Events:
\newline
[EVENTS]
\newline
\newline
Identify all causal pairs with first event causing the second one. You are responsible for looking into identifying the commonsense that is relevant to the given events and use them to identify whether there is a causal link or not. The missing commonsense can be some intermediate events that can help with identifying the link between the given pair of events.
\newline
\newline
Place your reasoning between <argument></argument> tags. 
Find any new causal relation for the given events. Please make sure that the newly added pairs do not violate the existing ones. Note that it is not possible to have both a causes b and b causes a. Also note that if a causes b and b causes c, then a causes c as well. Provide the ids of each pair of events with causal relation in <pair></pair> where the first event is the cause and the second is the effect and then place all pairs in separate lines in <pairs></pairs> tags.
    \newline
    \newline
    Example of output format:
    \newline
    <pairs>
    \newline
    <pair>1,2</pair>
    \newline
    <pair>3,4</pair>
    \newline
    </pairs>
    \\ 
 \bottomrule
\end{tabular}
\caption{Prompt used for agent with commonsense expertise.}
\label{tab:commonsense_prompt_single}
\end{table*}
\begin{table*}
\color{black}
\centering
\small
\begin{tabular}{@{}p{14cm}@{}}
\toprule
Given the following document:
\newline
[NEWS]
\newline
\newline
Events:
\newline
[EVENTS]
\newline
\newline
Identify all causal pairs with first event causing the second one.
\newline
1. looking into the temporality of the given events and evaluate the causality from the temporal perspective. Each causal pair must have a temporal relation such that the cause precedes the effect.
\newline
2. looking into common entities that are shared among the given events and evaluate the causality from that perspective. Events with causal relation can potentially be sharing some entities.
\newline
3. looking into the dependence of the given events and evaluate the causality from the dependence perspective. Events are dependent if removing one of them (assuming it did not happen) leads to making the other event irrelevant. 
\newline
4. looking into identifying the commonsense that is relevant to the given events and use them to identify whether there is a causal link or not. The missing commonsense can be some intermediate events that can help with identifying the link between the given pair of events.
\newline
\newline
Place your reasoning between <argument></argument> tags. 
Find any new causal relation for the given events. Please make sure that the newly added pairs do not violate the existing ones. Note that it is not possible to have both a causes b and b causes a. Also note that if a causes b and b causes c, then a causes c as well. Provide the ids of each pair of events with causal relation in <pair></pair> where the first event is the cause and the second is the effect and then place all pairs in separate lines in <pairs></pairs> tags.
    \newline
    \newline
    Example of output format:
    \newline
    <pairs>
    \newline
    <pair>1,2</pair>
    \newline
    <pair>3,4</pair>
    \newline
    </pairs>
    \\ 
 \bottomrule
\end{tabular}
\caption{Prompt used for agent with all 4 expertise.}
\label{tab:4agents_prompt_single}
\end{table*}
The prompts for agents with temporal, discourse, precondition and commonsense expertise are shown in Tables \ref{tab:temporal_prompt_single}, \ref{tab:discourse_prompt_single}, \ref{tab:precondition_prompt_single} and \ref{tab:commonsense_prompt_single} respectively.
Finally, we provided all the expertise to a single agent as shown in Table \ref{tab:4agents_prompt_single} to see how an agent equipped with all the necessary expertise can help with identifying causal relations.
The results are shown in Table \ref{tab:causal_graph_experts_ablations}.
\begin{table*}
\color{black}
\centering
\small
\begin{tabular}{@{}lcccc|cccc@{}}
\toprule
\multirow{2}{*}{Model} & \multicolumn{4}{c}{Graph-level} & \multicolumn{4}{c}{Pair-level} \\ \cmidrule{2-5} \cmidrule{6-9} 
                       & BAcc         & F1:C & F1:NC & Macro F1         & BAcc         &F1:C & F1:NC& Macro F1        \\ \midrule
Experts wo collab & 74.92 &  70.21 & 78.23 & 74.22 & 71.30 &  65.68 & 77.50 & 71.59 \\
Collab with experts & \textbf{79.27}  & \textbf{75.62} & \textbf{82.80}& \textbf{79.21} & \textbf{77.12} &  \textbf{72.96} & \textbf{82.11} & \textbf{77.51} \\
\midrule
Single agent with Temporal expertise &   76.50 & 69.59 & 81.97 & 75.78 & 68.65 & 59.72 & 74.34 &67.03
\\
Single agent with Discourse expertise &    75.02 & 69.74 & 79.71 & 74.73 & 69.04 & 60.52 & 74.53 &67.53
\\
Single agent with Precondition expertise &   74.35 & 68.67 & 79.08 & 73.87 & 69.14 & 61.27 & 74.35 & 67.81
\\
Single agent with Commonsense expertise&     72.37 & 66.89 & 77.14 & 72.12 &67.97 &59.20 &73.63 &66.42
\\ 
Single agent with 4 expertise &  75.21 & 69.35 & 80.15 & 74.87 & 69.14 &60.89 & 74.51 &67.70
\\
\bottomrule
\end{tabular}
\caption{Performance of single experts on causal graph generation using GPT-4o as the base LLM.}
\label{tab:causal_graph_experts_ablations}
\end{table*}
As can be seen from the results, adding the expertise explicitly to the prompt can help with identifying causal relations compared to the direct elicitation with the agent with all 4 expertise performing the best. However, there is still a large gap between the collaborative approach and any other settings.}

\subsubsection{Cost Analysis}\label{app:cost-analysis}
Computational costs are an important concern. Here we present an analysis of different settings. Table \ref{tab:cost_analysis} presents the number of LLM calls for different settings as well as the specific cost incurred for our studies.
\begin{table*}
    \centering
    \begin{tabular}{|l|c|c|c|}
    \hline
    \textbf{Model} & \textbf{LLM calls per context} & \textbf{Cost per context} & \textbf{Total cost (GPT-4o)} \\ \hline
    Direct         & 1                               & <0.01\$                   & \~1\$                        \\ \hline
    Pairwise       & n*(n-1)                         & \~0.04\$                  & \~4\$                        \\ \hline
    Debate         & m*r                             & \~0.35\$                  & \~30\$                       \\ \hline
    \end{tabular}
    \caption{Cost analysis of different causal graph generation approaches}
    \label{tab:cost_analysis}
\end{table*}
where n = number of events, m = number of experts and r = number of debate rounds. The total costs are for the entire test set. There is not much difference between direct and pairwise as the average number of events in the CRAB dataset is small. The difference would be higher for larger context sizes. Also, the debate costs more than pairwise, since debate uses longer contexts (previous agents' responses) which increases its cost. The estimate for the debate approach is under the assumption that all debates take for a fixed $r$ rounds, which might not be the case in practice. Experts might reach consensus earlier. An analysis of the outputs revealed that in 25\% of the cases, the conversation finishes early, thus reducing the overall costs. Reducing the computational cost of the debate approach is an important future work.

\subsubsection{Number of Experts Effect}\label{app:experts-number-analysis}
We measured the performance of our collaborative approach using different number of experts. As expected, adding more experts would result in better accuracy as shown in Table \ref{tab:number_of_experts}.
\begin{table*}
    \centering
    \begin{tabular}{l|cccc|ccccc}
        \toprule
        & \multicolumn{4}{c}{Graph-level} & \multicolumn{4}{c}{Pair-level} \\
        \cmidrule(lr){2-5} \cmidrule(lr){6-9}
        Number of experts & BAcc & F1 Pos & F1 Neg & Macro F1 & BAcc & F1 Pos & F1 Neg & Macro F1 \\
        \midrule
        1 expert & 76.50 & 69.59 & 81.97 & 75.78 & 68.65 & 59.72 & 74.34 & 67.03 \\
        3 experts & 78.88 & 74.68 & 82.81 & 78.85 & 74.33 & 69.36 & 80.05 & 74.71 \\
        4 experts & 79.27 & 75.62 & 82.80 & 79.21 & 77.12 & 72.96 & 82.11 & 77.51 \\
        \bottomrule
    \end{tabular}
    \caption{Effect of number of experts for causal graph generation.}
    \label{tab:number_of_experts}
\end{table*}

Also, one thing to note is that our agents are semantic relations experts that can help with identifying causal relations. We have identified 4 such experts that can help with identifying causal relations. Going beyond four diverse agents would require identifying additional experts that can help with causal link identification that is not explored in this work as we believe the current set of experts can capture all the subtleties needed to identify causal relations.

{\color{black}
\subsection{Collaboration Example}
\begin{table*}
\centering
\scriptsize
\begin{tabular}{@{}p{16cm}@{}}
\toprule
Article:
\newline
The US, and the technology sector which were among its biggest customers, face another Lehman Brothers moment with the failure of SVB. US regulators rushed to seize the assets of Silicon Valley Bank (SVB) on Friday after a run on the bank, the largest failure of a financial institution since the height of the financial crisis more than a decade ago. Silicon Valley, the country's 16th largest bank, failed after depositors - mostly technology workers and venture capital-backed companies - hurried to withdraw their money this week as anxiety over the bank’s situation spread. The bank could no longer cope with the massive withdrawals of its customers and its last attempts to raise new money did not succeed. US authorities therefore officially took possession of the bank and entrusted its management to the U.S. agency responsible for guaranteeing deposits, the Federal Deposit Insurance Corporation (FDIC). Little known to the general public, SVB had specialised in financing start-ups and had become one of the largest banks in the US by asset size: at the end of 2022, it had \$209 billion (€196 billion) in assets and about \$175.4 billion (€164.5 billion) in deposits. Its demise represents not only the largest bank failure since that of Washington Mutual in 2008, but also the second-largest failure of a retail bank in the United States. US Treasury Secretary Janet Yellen called several financial sector regulators together on Friday to discuss the situation, reminding them that she had "full confidence" in their ability to take appropriate action and that the banking sector remained "resilient".
\\
\midrule
Events:
\newline
1. Silicon Valley Bank (SVB) collapsed.
\newline
2. Depositors, mostly technology workers and venture capital-backed companies, hurried to withdraw their money this week.
\newline
3. SVB had specialised in financing start-ups.
\newline
4. The technology sector face crisis.
\\
\midrule
Conversation among agents (Round 1):
\newline
\newline
\textbf{Temporal agent:} Based on the temporal aspect of the events presented, I identify potential causal relationships. Specifically, the event in which depositors hurried to withdraw their money (Event 2) is directly linked to the collapse of Silicon Valley Bank (SVB) (Event 1). The withdrawal of funds indicates a panic reaction to the bank's instability, which leads logically to its failure. The sequence is clear: the withdrawals occurred before the bank's collapse. Thus, I find the causal pair: 2 causes 1. In examining other possible relations, I also look at the relationship between the technology sector facing a crisis (Event 4) and the collapse of SVB (Event 1). The document suggests that SVB's failure affects the entire technology sector, particularly given that many depositors were technology workers and companies reliant on venture capital. However, since Event 4 does not temporally precede Event 1, I cannot identify a direct cause-effect relationship there. The discourse agent's analysis also supports that the technology sector's crisis could stem from the bank's failure but does not establish a clear temporal sequence, which is crucial. Furthermore, the third event, regarding SVB’s specialization in financing startups (Event 3), lacks a direct causal link to the other events. Although it provides context for why the collapse is significant, it does not highlight a temporal or causal relationship with either the withdrawals or the bank’s failure. The disappearance of SVB affects the technology sector, but this does not suggest causation towards Event 4 without the preceding trigger of the bank's downfall.
\newline
\newline
\textbf{Discourse agent:} Upon reviewing the events presented, it is evident that there is a causal relationship between event 2 and event 1. Specifically, the action of depositors, mostly technology workers and venture capital-backed companies, hurrying to withdraw their money (event 2) directly contributed to the collapse of Silicon Valley Bank (SVB) (event 1). The shared entities here are the depositors, particularly those from the technology sector, which plays a critical role in understanding how one event leads to another. Additionally, event 4, which states that the technology sector faces a crisis, is a consequence of event 1's collapse of SVB, reinforcing that the technology sector's well-being is deeply tied to the stability of their financial institutions. This suggests a chain of causality could potentially exist linking these events, where the collapse of SVB (event 1) leads to the crisis in the technology sector (event 4). However, I find the identification of the causal relationship between event 3 (SVB’s specialization in financing start-ups) to be less direct. While SVB's specific business focus might have an indirect impact on the technology sector's crisis, the causal link is not as strong as the others because the collapse caused immediate consequences that directly affected depositors first. Thus, I propose the following causal pairs: - From event 2 to event 1 - From event 1 to event 4 The proposed causal pairs based on shared entities and their sequential relationships includes:
\newline
\newline
\textbf{Dependence agent:} Upon reviewing the events, I find that the second event, "Depositors, mostly technology workers and venture capital-backed companies, hurried to withdraw their money this week," is essential to the first event, "Silicon Valley Bank (SVB) collapsed." Without the hurried withdrawals of depositors, which were driven by anxiety over the bank's financial situation, the bank would not have experienced the severe liquidity crisis that ultimately led to its collapse. The dependencies between these two events are evident: if the depositors had not rushed to withdraw their funds, SVB might have remained solvent or at least had more time to address its financial discrepancies. Therefore, it is logical to conclude that the first event (SVB collapsed) is highly dependent on the second event (Depositors withdrawing funds). Further analysis from other agents also supports the idea that these events are temporally linked, as the withdrawals directly contributed to the bank’s collapse. In this case, I agree with my findings and believe these events can be causally paired. 
\newline
\newline
\textbf{Commonsense agent:} In reviewing the events presented, it is crucial to consider how commonsense reasoning plays a role in determining causal relationships. The first event states that "Silicon Valley Bank (SVB) collapsed" and the second event notes that "Depositors, mostly technology workers and venture capital-backed companies, hurried to withdraw their money this week." From a commonsense perspective, the collapse of a bank typically triggers panic among its customers, leading them to withdraw their funds to avoid losing their deposits. Therefore, it is reasonable to conclude that the collapse of SVB caused the depositors to rush to withdraw their money. This reasoning aligns with the context provided, as the document emphasizes the anxiety surrounding SVB's situation, further supporting the causal link between these events. Regarding the third event, "SVB had specialised in financing start-ups" this information does not directly contribute to the other events in terms of causality. It provides context but does not create a causal link to the bank's collapse or the subsequent actions of depositors. Finally, the fourth event mentions that "the technology sector faces crisis." The crisis in the technology sector appears to be a broader consequence of the bank's collapse rather than the immediate effect of the depositors' actions. However, it could be influenced by the bank’s collapse due to its role in financing startups within that sector. In conclusion, the pairs that I identify as having a causal link due to commonsense reasoning are: 1. The event that SVB collapsed caused depositors to hurriedly withdraw their money. 
    \\ 
 \bottomrule
\end{tabular}
\caption{Example of a collaboration among agents (Continued over the next two pages).}
\label{tab:collab-example}
\end{table*}
\begin{table*}
\centering
\scriptsize
\begin{tabular}{@{}p{16cm}@{}}
\toprule
Events:
\newline
1. Silicon Valley Bank (SVB) collapsed.
\newline
2. Depositors, mostly technology workers and venture capital-backed companies, hurried to withdraw their money this week.
\newline
3. SVB had specialised in financing start-ups.
\newline
4. The technology sector face crisis.
\\
\midrule
Conversation among agents (Round 2):
\newline
\newline
\textbf{Temporal agent:} In discussing the identified causal pairs, I maintain my stance on the relationship between the events involving the depositors and the collapse of Silicon Valley Bank (SVB). The sequence in which depositors hurried to withdraw their money (Event 2) directly precedes the bank's collapse (Event 1) suggests a clear causality based on the temporal aspect, supporting the idea that the withdrawals were a reaction to the perceived instability of the bank. Other agents have pointed out that the bank’s collapse could also be interpreted as causing the hurried withdrawals; however, from a temporal perspective, it is imperative to focus on the order of events as described. The actions taken by the depositors immediately following the realization of SVB’s precarious condition played a critical role in precipitating the bank’s failure. Furthermore, while there is a suggestion of causality between the bank’s failure and the technology sector facing a crisis (designated as Event 4), the timing indicates that this relationship is one of consequence rather than direct causation. The bank's collapse (Event 1) likely leads to the crisis the technology sector faces (Event 4), but it does not directly cause the struggles of the technology sector itself absent the preceding events. The dependence analysis aligns with my view that the second event (withdrawals) is critical to the first event (collapse), reinforcing the significance of the initial action that set off a chain reaction leading to SVB’s downfall. Hence, while I acknowledge the potential for a chain of causality, I caution against conflating that with direct causal links without a clear temporal precedent. In terms of new pairs, I observe that while the relationship established between Event 1 and Event 4 is tenuous, it does hold merit in suggesting that the effects of SVB's failure reverberate throughout the technology sector. Therefore, I suggest we include the pair representing that potential causality while keeping the temporal order intact. Thus, I propose the following pairs of events: 1. <pair>2,1</pair> - depositors hurried to withdraw their money causes the SVB to collapse. 2. <pair>1,4</pair> - SVB's collapse leads to the technology sector facing a crisis. The event relating to SVB's specialization in financing startups (Event 3) remains contextually relevant but does not provide a direct causal link to the failure or withdrawal actions, thus not necessitating a new pairing. In conclusion, I agree with the identification of the causal relationships present whilst emphasizing the necessity of maintaining the temporal order in these relationships. <pairs> <pair>2,1</pair> <pair>1,4</pair> </pairs>", 
\newline
\newline
\textbf{Discourse agent:} Upon reviewing the discussions, it seems we have a strong consensus regarding the causal link between event 2 (Depositors hurried to withdraw their money) and event 1 (Silicon Valley Bank collapsed). All agents acknowledge the shared entities, primarily the depositors, who are directly impacted by the bank's failure due to their financial ties. This shared aspect solidifies our understanding that the hurried withdrawals caused the bank's collapse. Additionally, the dependence analysis correctly emphasizes that if the depositors had not withdrawn their funds, the collapse may not have occurred, demonstrating a critical dependence. However, I notice a divergence in the interpretation of the relationships from event 1 (SVB collapsed) to event 4 (The technology sector faces crisis). One evaluator suggests that the crisis in the technology sector is a broader consequence of SVB's collapse, which is valid since many technology firms were reliant on SVB. They argue for a causal link, and I agree that the shared context linking SVB's downfall to the technology sector is relevant—without the failure of SVB, the financial uncertainty experienced by these firms may have been mitigated. Thus, we should maintain that relationship. Regarding the third event, there appears to be less agreement on its relevance to causality. While one evaluator argues that SVB’s specialization may not directly lead to causal inferences, the role of SVB in financing startups does connect it to the technology sector. However, this connection is more circumstantial than causal, and therefore I find it inappropriate to establish a causal relationship between event 3 and the others. In conclusion, I stand by our previously proposed causal pairs based on shared entities, circumstance, and their apparent dependencies: 1. From event 2 to event 1, as the depositors' actions directly led to the bank's collapse. 2. From event 1 to event 4, recognizing that SVB's failure significantly impacts the technology sector. Therefore, I propose we finalize our pairs as follows:", 
\newline
\newline
\textbf{Dependence agent:} After reviewing the previous discussions and analyses from each agent, I maintain my stance on the significance of the relationship between the events concerning the collapse of Silicon Valley Bank (SVB) and the actions of depositors. Firstly, the dependence agent's insights emphasize that the action of depositors withdrawing their funds directly contributed to the collapse of SVB. This aligns with my own analysis — removing the rush to withdraw funds would imply that the bank might have had the opportunity to stabilize itself, or at least regroup, which reinforces the notion that one event is indeed dependent on the other. Furthermore, the temporal agent identified that the withdrawals preceded the bank's collapse, clearly mapping out the timeline and reinforcing the causality — specifically, that the depositors' actions led to the bank's failure. To clarify, while the temporal agent and I agree on the causality direction from withdrawals to the bank collapse, it's essential to highlight the difference in wording: while I see the collapse being caused by the withdrawals, the temporal agent suggests the withdrawals cause the collapse. This distinction highlights the necessity to focus on the dependence clearly. In simplistically evaluating these relationships, I also acknowledge that there is a significant connection between the collapse of SVB and the ensuing crisis in the technology sector. Although the discourse agent drew a line from the collapse to the sector's crisis, I find it crucial to carefully assess the dependency factor. The technology sector’s crisis certainly appears to be a consequence of the bank collapsing; however, the timing here doesn’t demonstrate a justifiable sequence as it does with the withdrawals leading to the collapse. With respect to commonsense reasoning, the agent recognized that the typical reaction of depositors in a bank's collapse is to withdraw funds, logically supporting the correlation. This aligns with other evaluations, marking a consensus about this particular causal relationship. Upon reviewing the involved events, I propose the following new causal pairs, which stem from the previous evaluations: 1. The collapse of SVB (Event 1) does impact the crisis in the technology sector (Event 4) directly due to the shared entities and economic interdependence, meaning I can accept the discourse agent's identification of this as a causal pair without contradicting my earlier input. Therefore, my final assessment of causal pairs is:
\newline
\newline
\textbf{Commonsense agent:} In reviewing the proposed causal pairs, I find significant agreement with the conclusions drawn by the other agents regarding the relationship between Event 1 (SVB collapsed) and Event 2 (Depositors hurried to withdraw their money). The commonsense reasoning aligns quite well here; typically, when a bank is perceived to be unstable, panic ensues, prompting its depositors to withdraw their funds. This chain reaction is clearly articulated in the document, where it mentions that anxiety over the bank's situation spread among depositors. Therefore, I affirm the identification of the causal link from Event 1 to Event 2. However, I notice some discrepancies concerning Event 4, which states that "The technology sector faces crisis." While there is a sentiment shared by some agents that this event may be a direct consequence of Event 1, I believe that the connection is not as direct as they suggest. The technology sector's crisis feels more like a reverberation of the SVB’s collapse rather than an immediate outcome of the withdrawal actions by depositors. The relational path here needs some cautious interpretation as the crisis could also stem from other factors related to the banking situation beyond just SVB's singular issue. Additionally, while Event 3 details that "SVB had specialised in financing start-ups," this information only contextually positions SVB’s role in the broader financial ecosystem; it does not imply causality towards the collapse or the depositors' actions. It should be noted that while SVB's business focus significantly impacted its depositors, this aspect alone does not create a causal link to any of the critical events discussed. Given these evaluations, I would propose the following causal relation based on commonsense reasoning: 1. The event that SVB collapsed (Event 1) caused depositors to hurriedly withdraw their money (Event 2). 2. The action of depositors withdrawing their funds (Event 2) directly contributed to the collapse of SVB (Event 1). Thus, in summary, I suggest adding: - The causal pair 1,2 as it is defined. - The discourse of potential influences of the collapse on the tech sector does not provide a straightforward causal chain. With that being said, I will formalize the pairs in the requested format:", 
    \\ 
 \bottomrule
\end{tabular}
\caption*{Example of a collaboration among agents (Contd).}
\label{tab:collab-example2}
\end{table*}
\begin{table*}
\centering
\scriptsize
\begin{tabular}{@{}p{16cm}@{}}
\toprule
Events:
\newline
1. Silicon Valley Bank (SVB) collapsed.
\newline
2. Depositors, mostly technology workers and venture capital-backed companies, hurried to withdraw their money this week.
\newline
3. SVB had specialised in financing start-ups.
\newline
4. The technology sector face crisis.
\\
\midrule
Conversation among agents (Round 3):
\newline
\newline
\textbf{Temporal agent:} In reviewing the evaluations provided by the different agents, I find a strong consensus regarding the causal relationship between event 2 (Depositors hurried to withdraw their money) and event 1 (Silicon Valley Bank collapsed). Most agents highlight the crucial role of depositors' actions in precipitating the bank's failure, which aligns with my own analysis focusing on the temporal sequence of events where withdrawals occurred before the collapse, thereby establishing a clear cause-and-effect relation. However, there is some divergence concerning the relationship between event 1 (SVB collapsed) and event 4 (The technology sector faces crisis). Some agents suggest that the collapse leads to the crisis in the technology sector, which is valid considering the shared context and economic dependence between the bank and the firms within that sector. While I can appreciate this interpretation, I emphasize the importance of maintaining clarity regarding the temporal sequence and specificity of the causality being discussed. The crisis in the technology sector might indeed result from the SVB collapse but does not directly stem from the immediate events surrounding the withdrawals of depositors. Regarding event 3 (SVB had specialised in financing start-ups), I agree with the consensus that it does not establish a direct causal link with other events but offers contextual information about SVB's operational focus and its significance in relation to the tech sector. This context is key, but it does not contribute to a causative relationship in the same way that the withdrawals affect the bank's collapse. In light of this review, I support the pairs that have been identified so far: 1. From event 2 to event 1 (Depositors hurriedly withdrew their money causes SVB to collapse). 2. From event 1 to event 4 (SVB collapsed leads to the technology sector facing a crisis). These pairs maintain the temporal order that reinforces their causal relationship. I will formally present them in the specified format below:", 
\newline
\newline
\textbf{Discourse agent:} In reviewing the causal pairs identified by all agents, I've noted a significant agreement concerning the relationship between Event 2 (Depositors hurried to withdraw their money) and Event 1 (Silicon Valley Bank collapsed). The reasoning revolves around the shared entities, primarily the depositors from the technology sector, who acted in response to the anxiety surrounding SVB's financial instability. This shared context underscores the causal link proposed that the actions of those depositors led directly to the bank's failure. However, I’ve discerned a divergence in how the relationship between Event 1 and Event 4 (The technology sector faces crisis) is framed across evaluations. While there is a consensus that SVB's collapse impacts the technology sector, I lean towards understanding it as more of a reverberation rather than a direct causative action. One agent indicated that the crisis in the technology sector is indeed a consequence of SVB's failure, given the interdependence of SVB with technology companies whom they financed. I find merit in this argument—for instance, the failure of a key bank like SVB would likely destabilize economic conditions for its technology clients. Nevertheless, based on the temporal and dependence analyses as presented, it seems there exists a stronger causal chain from Event 2 to Event 1, where the withdrawal actions directly precipitate the bank’s collapse. In contrast, the relationship from Event 1 to Event 4, while consequential, doesn’t share the same immediacy or dependency, meaning that while SVB’s collapse may set the stage for a crisis in the technology sector, it is not directly causing it in the immediacy of causation that can be drawn from the first two events. Additionally, the notion that SVB’s specialization in financing startups (Event 3) does not yield a causal link appears to be widely supported across evaluations. It lends context to the discussion without providing direct causation. In conclusion, I therefore propose we retain our previously agreed relationships while clarifying the nature of the relationships based on temporal precedence and shared entities, leading us to establish the following causal pairs: 1. From Event 2 to Event 1, indicating that the hurried withdrawal actions caused the collapse. 2. From Event 1 to Event 4, acknowledging that the effects of SVB's failure may contribute to a broader crisis faced by the technology sector. Here are the pairs I propose: <pairs> <pair>2,1</pair> <pair>1,4</pair> </pairs>
\newline
\newline
\textbf{Dependence agent:} After reviewing the various analyses from my fellow agents, I would like to clarify my stance and engage in further discussion regarding the proposed causal pairs. Firstly, I agree with the identification of the causal pair where the action of depositors hurriedly withdrawing their money (Event 2) results in the collapse of Silicon Valley Bank (SVB) (Event 1). This aligns with my understanding of the dependence between these events, as without the influx of withdrawals, SVB might have had a chance to stabilize or mitigate the crisis. Removal of event 2 leads to a different narrative where the bank potentially continues operating, reinforcing the dependency demonstrated. However, I want to highlight a notable point of contention regarding the directionality of causation suggested by some evaluators concerning the pair linking SVB's collapse (Event 1) to the technology sector facing a crisis (Event 4). While I acknowledge the interconnectedness between SVB's failure and the struggles faced by the technology sector, as several deposits came from technology workers and firms, I think it's crucial to assess whether the technology sector's crisis is a direct consequence of the collapse or more of a reverberation of the bank's failure. The discourse surrounding the shared entities does indicate an intrinsic connection; however, without solid evidence of direct temporal dependence, it remains uncertain whether we can definitively say that SVB's collapse directly causes the tech sector crisis at this stage. Moreover, I concur with an observation made regarding Event 3, which states that SVB specialized in financing start-ups. While this provides valuable context, it does not establish a direct causal link to the other events, as it is purely informative. The focus should remain on the causative interactions demonstrated by Events 1 and 2, as they are more clearly defined. Therefore, I propose that we keep the causal relationship identified between Events 2 and 1 intact while maintaining a cautious perspective on the pairing between Events 1 and 4. Although it certainly appears that SVB's collapse impacts the technology sector, I argue that we need a clearer causal framework for this association if we are to formalize it without undermining the integrity of the established pairs. Given my reiterated stance on the need for assessment of direct causality based on dependence, I suggest the following pairs as valid causal relations: <pairs> <pair>2,1</pair> <pair>1,4</pair> </pairs>
\newline
\newline
\textbf{Commonsense agent:} Upon reviewing the evaluations from different agents regarding the causal pairs identified in the given events, I align myself strongly with the conclusions drawn about the relationship between Event 1 (Silicon Valley Bank collapsed) and Event 2 (Depositors hurried to withdraw their money). The reasoning across the evaluations supports the idea that the bank's collapse instigated a panic response from depositors, which is a common-sense understanding of financial crises. Without the collapse, there would be no immediate incentive for depositors to withdraw their funds, thereby establishing a causal link from Event 1 to Event 2. However, there is some divergence when it comes to the causal relationship between Event 1 (SVB collapsed) and Event 4 (The technology sector faces crisis). While it's acknowledged that the collapse of SVB likely contributes to a broader crisis in the technology sector, this connection appears to be more of a long-term consequence rather than a direct causation from the collapse to the crisis. I believe this point is emphasized by the discussions that note the timing and context of these events. Therefore, I support the identification of the causal pair between Event 1 and Event 2 but remain cautious about establishing a direct link between Event 1 and Event 4 without considering additional contextual factors that may impact the technology sector. Additionally, while one agent argued strongly for the link from Event 2 back to Event 1, viewing the depositors' actions as critical in leading to the collapse, I maintain that the foundational causality flows from the collapse to the withdrawals, rather than the other way around. The panic-induced withdrawals are indeed a reaction to the perceived instability created by the collapse. So, my argument is focused on maintaining that Event 1 causes Event 2 directly, while the link between Event 1 and Event 4, although valid, feels more consequential and less immediate based on the commonsense understanding of financial operations. Thus, I reaffirm the following pairs based on the assessments: 1. <pair>1,2</pair> - SVB collapsed causes depositors to hurriedly withdraw their money. 2. <pair>1,4</pair> - SVB's collapse leads to the technology sector facing a crisis. However, I do not propose a direct causal pair that contradicts already established relations. In conclusion, I stand by my stance of identifying the core causal pairs as described and want to highlight that the nature of the relationship from Event 1 to Event 4 is critical to explore, but might require careful interpretation to avoid drawing unfounded conclusions about immediate causation without sufficient evidence of direct linkages between each response and outcome in the timelines presented. Therefore, I finalize my assessment of causal pairs as follows:<pairs> <pair>1,2</pair> <pair>1,4</pair> </pairs>
    \\ 
 \bottomrule
\end{tabular}
\caption*{Example of a collaboration among agents (Contd).}
\label{tab:collab-example3}
\end{table*}
A full history of a collaboration across experts are shown in Table \ref{tab:collab-example}. In this example, the majority of experts except for the commonsense expert agree on a causal link from event 2 to 1 however the ground-truth causal link is in-line with what the commonsense expert believes is true. This is an example of a single expert failure to persuade the other experts even though it is making the right decision.}

\section{Causally Explainable Event Likelihood Details}\label{appendix:eel}

\paragraph{\textbf{1) Event Extraction}} Given the input text $D$ and the query event $e_q$, the first step is to extract all the events $E = <e_1,e_2,...e_n>$ mentioned in the text. In an in-context learning setup, given $D$ and an extraction prompt $p_e$, we use model $\cal{M}$ to generate a list of all events that are mentioned in the news article:
\begin{equation}
 E = M(D; p_e)
\end{equation}
The extracted events $E$ will then be used as explanation units to describe other events that are not mentioned in the text.

\paragraph{\textbf{2) Causal Graph Generation}} The next step creates a causal graph $G(E,R)$ from the extracted events $E$ as the nodes 
and edges are causal relations $R$, where $(e_i, e_j) \in R$ implies that $e_i$ is the cause of the effect $e_j$ using our proposed collaborative approach. We use our proposed collaborative causal graph generation approach to generate the causal graph of events.

\paragraph{3) Event placement}
To assess whether a query event $e_q$ is likely in the context of the given events, we need to identify its potential causes and effects in the causal graph. This translates to finding an insertion point within the causal graph. If the query event cannot be inserted, then the event is not likely. If the query event can be inserted, then the model has found causal connections that support its likelihood in the scenario.
To insert the query event, we provide the model $\cal{M}$ with the text, causal graph, and a new prompt $p_p$ to encourage it to alter the graph to include $e_q$.

\begin{equation}
 R^{\prime} = M(D; E; R; e_q; p_{p})
\end{equation}
Prompt $p_p$ is set up to answer two questions: (1) what are the events that are caused by $e_q$? and (2) what are the events that can cause $e_q$? (For the full set of prompts please refer to Figure \ref{fig:prompts} in Appendix \ref{prompts}).
The responses to these two questions modify the original document graph $G(E,R)$ and create a new graph $G(E^{\prime},R^{\prime})$ in which $E^{\prime}$ includes all events $E$ plus $e_q$ and $R^{\prime}$ consists of all relations between events in $E^{\prime}$.

\paragraph{3) Explanation generation}
If $e_q$ is placed in graph $G(E^{\prime},R^{\prime})$, it means that the event is likely and then any path $P$ in the graph that includes $e_q$ can be seen as a causal explanation for its likelihood, as the path encompasses all previous causes and subsequent effects of $e_q$. 


\begin{figure*}
    \centering
    \includegraphics[width=0.95\linewidth]{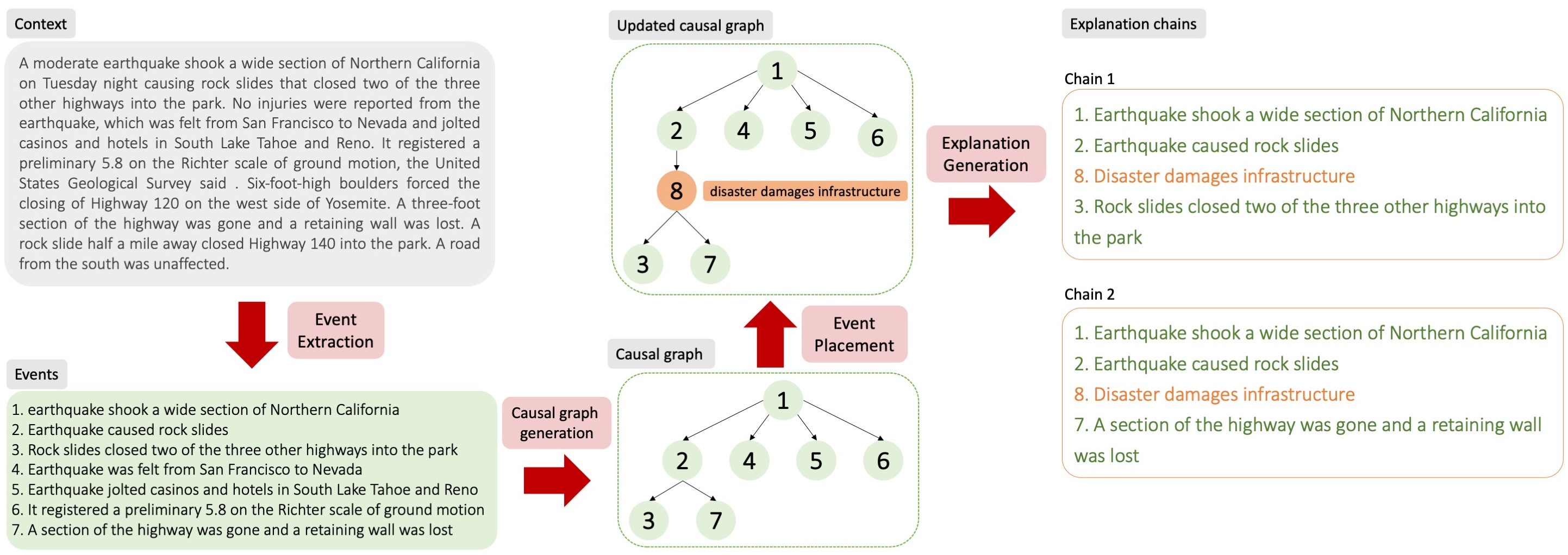}
    \caption{Overview of causal graph-based event likelihood prediction approach.}
    \label{fig:overview}
\end{figure*}
\subsection{Models Details}\label{app:methods}
We use a decoder-only instruction-following API-based model, GPT-4 \cite{openai2023gpt4}, as the base LLM for our experiments. The following sections describe all the models used in our evaluations.
The prompts we use for each step can be find in the Appendix \ref{prompts}.

\paragraph{Graph-based causal chain}
Our main contribution, Causally Explainable Event Likelihood \method\ captures explicit causal relations between events by creating its causal graph, and using it for likelihood prediction and later explanation. See Section \ref{sec:celp}.

Depending on where an event is placed, we can have multiple paths as explanations but we only choose one as the final event likelihood explanation. We can use different heuristics to pick a chain that contains the query event as the final explanation. We experimented with two: (a) select a random path and (b) select the longest path. 

\paragraph{One-shot causal chain}
In this baseline setting, we first provide GPT-4 with the in-context examples that include the text and causal chains (similar to the outputs of our graph-based approach) and then prompt it to directly generate causal chains containing the query event (prompts in Appendix \ref{prompts}). Similar to a graph-based causal chain approach, we would have multiple output chains, and use the same random path and longest path heuristics for picking the final explanation.

\paragraph{One-shot causal chain with events}
One of the main steps of our approach is extracting events that have already been described in the given news article. 
We therefore use another baseline setting, where GPT-4 is also given the list of events in the article and is then asked to generate the chains similar to the previous one-shot baseline. 
We use the same set of events that are extracted using GPT-4 for \method (see prompts in Appendix \ref{prompts}).

\paragraph{One-shot causal chain with reflection}
The reflection technique enables the model to analyze its own mistakes and improve its performance \cite{shinn2023reflexion}. In this setting we first generate causal chains and then ask the model to evaluate its previously generated chains for causality and check whether the events in the chain are causally related and if not, the model is asked to update its previously generated responses to better reflect the causal relations (refer to prompts in Appendix \ref{prompts}).

\subsection{Dataset Curation and Statistics}\label{append:data}
The task requires generating a likelihood explanation for an event, given a news article as the context. Therefore, the data needs to be in form of \textit{(news article, query event)} pairs. The \textit{query event} is an event which is relevant to the underlying scenario but not directly mentioned in the text.

We curated a small test set using the Annotated NYT corpus \cite{sandhaus2008new} with 520 \textit{(news article, query event)} pairs. More details on the curation process and the data distribution are available in the Appendix \ref{append:data}.

To generate such instances, we first select 80 news articles from the Annotated NYT corpus \cite{sandhaus2008new} corresponding to the high-level domains of natural disaster, IED, kidnapping, financial crimes, and disease outbreaks. These were chosen to correspond with existing manually-curated schemas \cite{devare2023sageviz} that contain the relevant events that typically occur in these five domains.
Then, for each article, we randomly sample 5-7 events from the curated schema that corresponds to the article's domain and create the \textit{(news article, query event)} pairs. This process results in a test set of 520 pairs.

Each event in the test set can be one of the following: likely to have occurred only in the past, likely to occur only in the future, likely to have both already occurred and occur in the future (for example, an event like `authorities investigate the attack'), or unlikely (when there is a contradiction or lack of evidence). The distribution of events based on their likelihood is shown in Table \ref{tab:data-dist} in Appendix \ref{append:data}.

We analyzed the instances in the created test set and for each pair evaluate whether the query event is likely/unlikely to have occurred in the past or likely/unlikely to occur in future with respect to the given context. We the compute the percentage of each category and report the numbers in Table \ref{tab:data-dist}.
\begin{table}
\small
\centering
\begin{tabular}{ccc}
\toprule
\multicolumn{1}{c}{} & \multicolumn{1}{c}{Likely} & \multicolumn{1}{c}{Unlikely} \\ \midrule
Past                 &           53\%                 &         47\%                     \\
Future               &            50\%                &           50\%                   \\ \bottomrule
\end{tabular}
\caption{Distribution of likely/unlikely events over time.}
\label{tab:data-dist}
\end{table}

\subsection{Evaluation Criteria}\label{app:criteria}
The generated explanations need to be based on evidence from the text and free of hallucinations.

Unfortunately, there are no available datasets with gold explanation chains that we could have used for evaluating our generated explanations. Therefore, to evaluate the quality of the generated explanations for the novel task of explainable likelihood prediction, we have two versions of the evaluation: GPT-based and human-based on a subset of all results to verify the GPT-based evaluation. 
Human evaluation is more trustworthy, however, the turn-around time and the costs associated with it convinced us to do a GPT-based evaluation for our full test set (again only for the explainable likelihood prediction task). However, to verify the GPT-4 results we also did a human evaluation on a subset of the data and observed similar trends. Human evaluation is also not done by ourselves (authors of the paper), but with two external annotators.

The different parts to an event's explanation should be coherent and causally related, and the explanations should be as informative as possible by connecting all pieces of available evidence together. 
Both the graph-based explanation and one-shot explanation generation approaches are instructed to use causality to glue different events together. Therefore, our evaluation approach also uses causality as the primitive question to measure for the desired criteria described earlier.

We use the main evaluation question, \textit{Can event ``EVENT1'' cause event ``EVENT2''?} to directly evaluate each causal relation between event pairs. For a given explanation chain $p \in P$ of length $n$ and for all $(e_i,e_{i+1})$ in $p$, we ask this primitive question to identify all causal relations. We record $n-1$ responses for each chain of size $n$, and based on the answers, we introduce three metrics: \textbf{\textit{Causality}}, \textbf{\textit{Informativeness}}, and \textbf{\textit{Coherence}}.
These metrics allow for comparison across systems and are defined in detail in the Appendix \ref{app:criteria}.

\paragraph{1. Causality}
The causality metric measures how good the generated explanations are in terms of capturing the cause and effect relations between events that are generated as part of the explanation chain, and might be relevant to the query event $e_q$.
Using the responses from the primitive question, we compute the percentage of correct causal relations (number of correct causal relations/total number of event pairs in the chain). 

\paragraph{2. Informativeness}
To measure informativeness, we first locate $e_q$ in the explanation chain. Then, we move in both directions and count the number of correct causal relations until an incorrect relation is met. The resulting chain is the longest chain with consecutive positive causal links including $e_q$.
The length of the longest sub-chain is the informativeness score. 

\paragraph{3. Coherence}
A coherent system is capable of correctly connecting pieces of information based on causal relations to form a clear likelihood explanation. 
Similar to informativeness, we find the longest sub-chain in which all event pairs are correctly causally connected (based on responses to the primitive causal question). However, this measure ignores the position of $e_q$ to find the longest sub-chain.
The length of this longest sub-chain is the coherence score.

We use three metrics based on the pairwise causality comparison because they capture the different subtleties of explanations.
Causality gives the overall chain accuracy between event pairs, but it lacks perspective on explaining $e_q$.

Informativeness measures specific aspects of how the chain interacts with $e_q$ and coherence cares for long range dependencies.
Suppose a chain $(e_1,e_2,e_3,e_q,e_4,e_5,e_6)$ with corresponding answers to the primitive question $[1,1,0,0,1,1]$. This chain has a Causality score of $0.67$, but an Informativeness score of $0$ and a Coherence score of $2$. 
This shows that even though the system correctly captured more than half of the casual relations, it completely fails in informativeness with respect to the event we care about ($e_q$).

{\color{black}
\subsubsection{A note on the choice of evaluation metrics}
It is worth mentioning to note that there does not exist a dataset which consists of (articles, query event, likelihood label), we created our own test set as described in Section \ref{eel_exp} and Appendix \ref{append:data}.

We conducted an early study to elicit the likelihood labels from human annotators but by analyzing the responses, and as stated in \cite{romanou2023crab}, human perception of causality usually depends on background context, implicit biases, epistemic state, and lack of information, making the task of actual causality attribution challenging. We also observed similar trends when conducting our preliminary annotation. We observed that the subjectivity of the human annotation would result in diverse responses as different annotators might have interpreted the given context differently, resulting in a large disagreement. Low inter-annotator agreement hinders the quality of the data therefore we opted out of using such annotations. As an alternative, we added explanation chains and observed, when having a clear explanation chain attached to a query event, the likelihood prediction task seemed to be a much easier task for human annotators with higher agreement.}

\subsection{Prompts}\label{prompts}
\begin{figure*}
    \centering
    \includegraphics[width=1\linewidth]{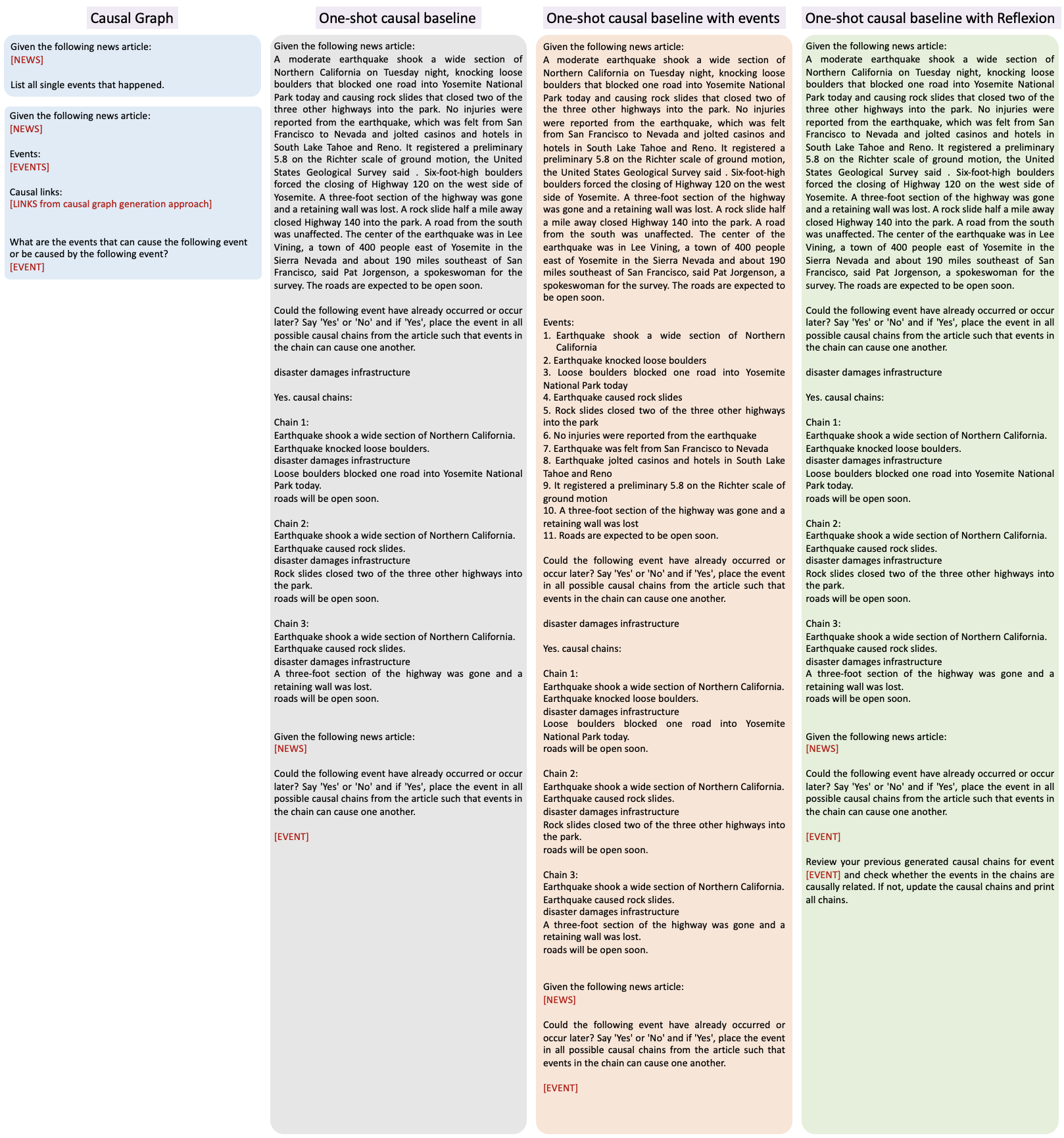}
    \caption{All the prompts used for models evaluated for event reasoning tasks. The \textcolor{red}{red} placeholders are reserved for individual instances.}
    \label{fig:prompts}
\end{figure*}

We have used different sets of prompts for the methods we have used throughout the paper. Figure \ref{fig:prompts} shows the full set of prompts for different system.

\subsection{More Results and Analysis}\label{more_results}
\subsubsection{GPT-4 evaluation correlation with humans}
There are no available datasets with gold explanation chains that we could have used for evaluating our generated explanations. Therefore, to evaluate the quality of the generated explanations for the novel task of explainable likelihood prediction, we have two versions of the evaluation: GPT-based and human-based on a subset of all results to verify the GPT-based evaluation. Human evaluation is more trustworthy, however, the turn-around time and the costs associated with it convinced us to do a GPT-based evaluation for our full test set (again only for the explainable likelihood prediction task). However, to verify the GPT-4 results we also did a human evaluation on a subset of the data and observed similar trends. Human evaluation is also not done by ourselves (authors of the paper), but with two external annotators.

We use GPT-4 for primitive questions since the turn-around time is faster and it requires less communication.
To verify that the results from GPT-4 are reliable, we selected all test instances from 10 articles and provided human annotators (students) with their explanations. Human annotators are asked the same primitive evaluation question \textit{Can event ``EVENT1'' cause event ``EVENT2''?}. 
We compare their responses to GPT-4 in Table \ref{tab:main-table-human}.
The human scores are generally within 10\% of GPT-4 for all metrics, giving us confidence to use GPT-4 for the pairwise causal question.

\begin{table*}
\centering
\begin{tabular}{l|lll|lll|lll}
\multirow{2}{*}{}             & \multicolumn{3}{c|}{\textbf{Causality}} & \multicolumn{3}{c|}{\textbf{Informativeness}} & \multicolumn{3}{c}{\textbf{Coherence}} \\
                              & \multicolumn{1}{c}{\small{Baseline}} & \multicolumn{1}{c}{\small{\method}} & \multicolumn{1}{c|}{\small{Tie}} & \multicolumn{1}{c}{\small{Baseline}} & \multicolumn{1}{c}{\small{\method}} & \multicolumn{1}{c|}{\small{Tie}} & \multicolumn{1}{c}{\small{Baseline}} & \multicolumn{1}{c}{\small{\method}} & \multicolumn{1}{c}{\small{Tie}} \\
\hline
human eval  & 27.14& 37.14 & 35.71 &  34.74 &  44.21 & 21.05 & 32.86 &37.14 &30.00\\
GPT-4 eval &  21.12 &  41.58 & 37.29 &  31.92 &  48.44 & 19.64 & 30.03 &36.96 &33.03\\
                       
\end{tabular}
\caption{Comparison of human and GPT-4 evaluation of the best-performing baseline and \method}
\label{tab:main-table-human}
\vspace{-3mm}
\end{table*}

\subsubsection{Selection Heuristics Comparison}\label{append:heuristics}
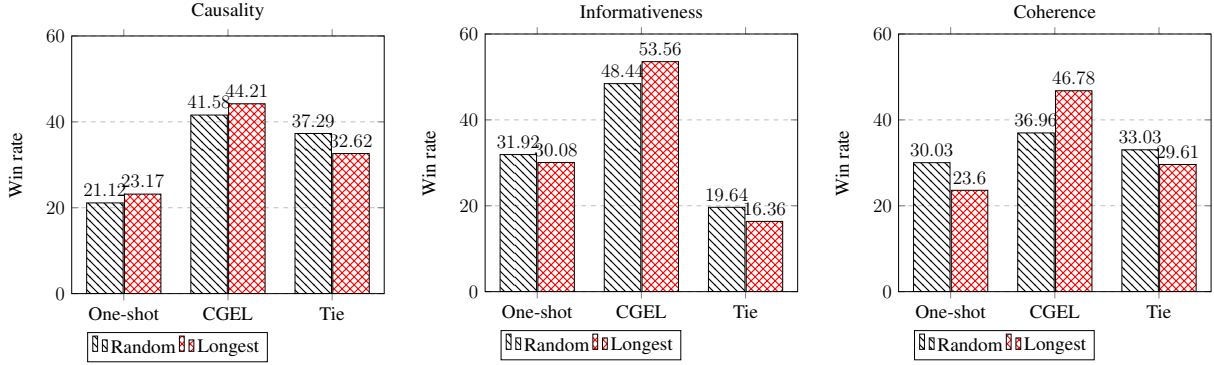
\begin{figure*}
\large
\begin{minipage}{0.32\textwidth} \centering
\begin{tikzpicture}[scale=0.6]
    \begin{axis}[
       ybar=2*\pgflinewidth,
        bar width=0.8cm,
        ymajorgrids = true,
        grid style=dashed,
        title=Causality,
        ylabel={Win rate},
        nodes near coords,
        symbolic x coords={One-shot,\method,Tie},
        xtick = data,
        scaled y ticks = false,
        enlarge x limits=0.25,
        ymin=0,
        ymax=60,
        legend columns=2,
        legend style={at={(0.05,-0.15)},anchor=north west},
    ]

    \addplot[style={black,pattern color=black,pattern = north west lines}]
    coordinates {(One-shot,21.12)(\method,41.58)(Tie,37.29)};
    
    \addplot[style={black,pattern color=red,pattern = crosshatch}]
    coordinates {(One-shot,23.17)(\method,44.21)(Tie,32.62)};

\addlegendentry{Random}
\addlegendentry{Longest}

    \end{axis}
\end{tikzpicture}
\end{minipage}\hfill
\begin{minipage}{0.32\textwidth} \centering
\begin{tikzpicture}[scale=0.6]
    \begin{axis}[
       ybar=2*\pgflinewidth,
        bar width=0.8cm,
        ymajorgrids = true,
        grid style=dashed,
        title=Informativeness,
        ylabel={Win rate},
        nodes near coords,
        symbolic x coords={One-shot,\method,Tie},
        xtick = data,
        scaled y ticks = false,
        enlarge x limits=0.25,
        ymin=0,
        ymax=60,
        legend columns=2,
        legend style={at={(0.05,-0.15)},anchor=north west},
    ]

    \addplot[style={black,pattern color=black,pattern = north west lines}]
    coordinates {(One-shot,31.92)(\method,48.44)(Tie,19.64)};
    
    \addplot[style={black,pattern color=red,pattern = crosshatch}]
    coordinates {(One-shot,30.08)(\method,53.56)(Tie,16.36)};

\addlegendentry{Random}
\addlegendentry{Longest}

    \end{axis}
\end{tikzpicture}
\end{minipage}\hfill
\begin{minipage}{0.32\textwidth} \centering
\begin{tikzpicture}[scale=0.6]
    \begin{axis}[
       ybar=2*\pgflinewidth,
        bar width=0.8cm,
        ymajorgrids = true,
        grid style=dashed,
        title=Coherence,
        ylabel={Win rate},
        nodes near coords,
        symbolic x coords={One-shot,\method,Tie},
        xtick = data,
        scaled y ticks = false,
        enlarge x limits=0.25,
        ymin=0,
        ymax=60,
        legend columns=2,
        legend style={at={(0.05,-0.15)},anchor=north west},
    ]

    \addplot[style={black,pattern color=black,pattern = north west lines}]
    coordinates {(One-shot,30.03)(\method,36.96)(Tie,33.03)};
    
    \addplot[style={black,pattern color=red,pattern = crosshatch}]
    coordinates {(One-shot,23.60)(\method,46.78)(Tie,29.61)};

\addlegendentry{Random}
\addlegendentry{Longest}

    \end{axis}
\end{tikzpicture}
\end{minipage}\hfill
\caption{Performance of systems based on different selection heuristics. Each bar shows the difference in performance of selection criteria on the specified metric.}
  \label{fig:heuristics-new}
\end{figure*}

When a model generates its likelihood explanation, it usually results in more than one explanation chain. 
In our graph-based approach, this occurs when an event is placed so that the graph contains multiple paths crossing it.
The one-shot baselines are actually instructed to do so to have similar output chains as the \method.
Because of this, we compared several heuristics to select the one best chain.
To compare the two common heuristics of random path and longest path, we compared both with the one-shot baseline against our graph-based \method\.
The results are shown in Figure~\ref{fig:heuristics-new}. As can be seen, regardless of the selection strategy, the results are similar. \method\ outperforms the one-shot baseline, and for longer chains, the gap is even larger.

We evaluate different heuristics select the final chain that is provided as the explanation in Figure \ref{fig:heuristics-new}. Each bar shows the difference in performance of selection criteria on the specified metric. The gap is very small for the causality whereas the longer chains lead to larger gap between the \method and the baseline in terms of informativeness and coherence. However, regardless of the selection heuristic, \method outperforms the baseline in all 3 dimensions.

\subsubsection{Error Analysis: Graph Structure}

\begin{table}
\centering
\small
\begin{tabular}{l|ccc}
\toprule
\multicolumn{1}{l}{Type} & root & non-leaf & leaf  \\ 
\midrule
Low-causality & 25.5\% &  46.8\% & 27.7\%    \\
Low-informativeness &  26.9\% &     32.9\%      &   40.2\%     \\
Low-coherence &   25.4\%  &        40.3\%        & 34.3\% \\
\midrule
Total \% of query node type & 25\% & 36\% & 39\% \\
\bottomrule
\end{tabular}
\caption{Error distribution over different node types. Each row represents the \% of error for a specific type when \method is worse than baseline.}
\label{tab:error-analysis}
\vspace{-5mm}
\end{table}
The query event can be placed as a root, non-leaf, or leaf node in the causal graph. 
We want to study how placement effects accuracy.
The last row in Table \ref{tab:error-analysis} shows the distribution of these locations across all test graphs.
When \method is worse than the baseline for any of the metrics, we report the percent of times it was each type of node as shown in Table \ref{tab:error-analysis}.
As seen from Table \ref{tab:error-analysis}, non-leaf nodes have higher error rates on average as it is more difficult for the system to correctly identify the relations of a query event with both the potential parents and children. One possible explanation for this is that the system has to identify the correct causal links with both the children as well as the parents of the event in question and this seems to be a more difficult task to do. Capturing longer-range dependencies is also a challenging task for the system to do as shown by higher error rate compared to the other two metrics, informativeness and coherence regardless of the node type. 
Leaf nodes are in the second position, as capturing longer dependencies is also a challenging task especially for deeper graphs.

\subsection{Sample Explanations}\label{apend:examples}
The \method generates explanations if an event is placed in the graph. Table \ref{tab:exp-examples} lists a few instances of the generated explanations and whether they are correct or not and the reason as to why there is an error in the generated explanation.
\begin{table*}
\scriptsize
\centering
\begin{tabular}{p{2.5cm}|p{8cm}|p{3.5cm}}
\toprule
Event & Explanation & Note\\
\midrule
disaster hurts people & An earthquake with a magnitude of 5.3 hit the western province of Kermanshah in Iran on Tuesday night.
Two milder aftershocks rocked the area.
disaster hurts people. The earthquake injured 15 people.
 & Correct, placed as a non-leaf node
\\
\midrule
rescue teams travel & The death toll from the earthquake on Saturday rose to about 5,000, with more than 15,000 injured and 100,000 displaced. Emergency teams previously deployed in Yogyakarta for the expected eruption of Mount Merapi shifted their resources to Bantul. A command center for the rescue effort was established in Bantul. rescue teams travel. Significant amounts of aid began arriving in Bantul on Monday. The convoy of trucks caused heavy traffic along the road to Bantul, further delaying aid & Correct, placed as a non-leaf node
\\ 
\midrule
government sends equipment & A series of monster tornadoes hit two states. The tornadoes destroyed more than 1,500 buildings. government sends equipment. & Correct, placed as a leaf node\\
\midrule
people hear evacuation order & Heavy rains over the weekend soaked an area from San Antonio to Austin, and from to Houston and Victoria. Persistent rain caused flooding in Texas, leading to the death of 15 people. Residents in Cuero were being urged to evacuate or pay close attention to the Guadalupe River, which was already three feet over the record flood level. people hear evacuation order. About 7,000 people were evacuated, with 5,300 of those seeking refuge in shelters in 51 locations.& Correct, placed as a non-leaf node \\
\midrule
\midrule
explosion kills people & A suicide bomber struck at a crowded market in the southeastern province of Paktia, killing at least 10 people and wounding 32. explosion kills people. The victims of the Paktia explosion were mainly shopkeepers and laborers waiting to be hired for daily work. & Incorrect, mixing correlation with causation\\
\midrule
disaster damages infrastructure & Heavy rains over the weekend soaked an area from San Antonio to Austin, and from to Houston and Victoria. Persistent rain caused flooding in Texas, leading to the death of 15 people. Residents in Cuero were being urged to evacuate or pay close attention to the Guadalupe River, which was already three feet over the record flood level. disaster damages infrastructure. The rain hampered efforts to clean up the damage and assess its cost.& Incorrect, incorrect causal relations with respect to the query event\\
\midrule
scientists invent drug & An outbreak of a rare but potentially blinding eye infection occurred, leading to the withdrawal of a contact lens cleaning solution by its manufacturer. The outbreak has involved at least 138 patients. Investigations of the acanthamoeba outbreak were initiated by the disease control agency, the Food and Drug Administration, state and local health departments, and eye doctors around the country. Investigators are exploring a variety of possibilities in seeking the cause of the outbreak, including changes in chlorination practice, the chemicals in the AMO solution, and the properties of newer generation silicone lenses. scientists invent drug& Incorrect. Event representation that consists of more than one concept, for example the first event has both outbreak and withdrawal, therefore leading to incorrect causal link with respect to the next event.\\
     
     \bottomrule
\end{tabular}
\caption{Sample correct and incorrect explanations generated for the test instances by \method.}
\label{tab:exp-examples}
\end{table*}

\section{Event Forecasting}\label{app:forecast}

\subsection{Data Processing}\label{app:forecast:data}
The ForecastQA dataset from \citet{jin2021forecastqa} consists of two main categories of questions; yes-no questions and multi-choice questions. We only consider the yes-no section of the questions for this evaluation as this type of questions can be more easily converted to the event likelihood question (a yes-no question can be considered as whether an event is likely or not).
Each question has an associated set of $10$ retrieved articles, which were used to generate the context for an \textit{open-book} forecasting setting. Since the concatenation of all $10$ articles is too large to fit as context, we follow the same summarization procedure used by \citet{jin2021forecastqa}. We use an off-the-shelf extractive summarizer (the trained DistillRoberta \cite{Sanh2019DistilBERTAD} checkpoint{\tiny \footnote{\url{transformersum.readthedocs.io/en/latest/}}}) to first summarize each article to $2$ sentences and then concatenate them to form the context news article.

\subsection{Explanations Generated for ForecastQA}
Applying \method to forecasting task has the additional benefit of generating explanations as to why system thinks an event is likely to occur in future. Table \ref{tab:forecast-examples} shows some of such explanations as the byproduct of the forecasting task.
\begin{table*}
\scriptsize
\centering
\begin{tabular}{p{3.5cm}|p{0.7cm}|p{10.3cm}}
\toprule
Question & Answer & Explanation   \\ \midrule
      Will Trump order agencies to assist William Barr to review the Russian probe in May 2019?",    &    \centering{1}    &   1. Barr told lawmakers that he will be looking to the \"genesis\" of the the FBI's counterintelligence investigation into potential ties between members of Donald Trump's presidential campaign and the Russian government began in 2016. 
2. The news of Barr's investigation was likely viewed as a welcome development to the President, who has regularly called for an investigation into the origins of the Russia probe.
3. Trump will order agencies to assist William Barr to review the Russian probe in May 2019.
        \\
        \midrule
     Is the attrition rate of rap cases going through the justice system causing mounting alarm, with campaign groups warning that it threatens to create a culture of impunity  In September 2019? &     \centering{1}    &     1. The budget for the Crown Prosecution Service in England and Wales was reported to be at least 30 per cent lower than it was in 2010. 2. Lawyers dismissed the chancellor's pledge to boost spending on the criminal justice system as an insignificant increase. 3.In September 2019, the attrition rate of rap cases going through the justice system is causing mounting alarm, with campaign groups warning that it threatens to create a culture of impunity.         \\ 
     \midrule
     Will Just Eat be bought by Takeaway.com in July 2019? &\centering{1}  & 1. Alex Captain, Cat Rock's founder and managing partner, spoke with many other shareholders who share his conclusion that a fair merger is a better alternative than relying on the board to choose a new leader for the company.2. The Amsterdam-headquartered food delivery company Takeaway.com struck a deal to buy the German food delivery operations from Delivery Hero. 3.Just Eat will be bought by Takeaway.com in July 2019.\\
     
     \bottomrule
\end{tabular}
\caption{Sample generated explanations for the ForecastQA instances that are predicted as likely by \method.}
\label{tab:forecast-examples}
\end{table*}
\end{document}